\documentclass[lettersize,journal]{IEEEtran}
\usepackage{amsmath,amsfonts}
\usepackage{algorithmic}
\usepackage{algorithm}
\usepackage{array}
\usepackage[caption=false,font=normalsize,labelfont=sf,textfont=sf]{subfig}
\usepackage{textcomp}
\usepackage{stfloats}
\usepackage{url}
\usepackage{verbatim}
\usepackage{graphicx}
\usepackage{cite}
\usepackage{dsfont}
\usepackage{multirow}
\usepackage{multicol}
\usepackage{tabularx}
\usepackage{booktabs}

\usepackage{overpic}
\usepackage{color}
\usepackage{makecell}
\newcommand{\topone}[1]{\textbf{\textcolor{black}{#1}}}
\newcommand{\toptwo}[1]{\textbf{\textcolor{black}{#1}}}

\definecolor{dg}{rgb}{0,0.694,0.298}
\definecolor{purple}{rgb}{0.4,0.176,0.569}
\definecolor{royalblue}{RGB}{65,105,225}
\usepackage{pifont}
\newcommand{\figref}[1]{Fig.~\ref{#1}}
\newcommand{\reqref}[1]{Eq.~\eqref{#1}}
\newcommand{\secref}[1]{Sec.~\ref{#1}}
\newcommand{\tableref}[1]{Table~\ref{#1}}

\newcommand{\yu}{\textcolor[rgb]{0.0, 0.0, 0.0}}
\newcommand{\qing}{\textcolor[rgb]{0.0, 0.0, 0.0}}

\newcommand{\kk}{\textcolor[rgb]{0.0,0.0,0.0}}
\newcommand{\ryn}{\textcolor[rgb]{0,0,0.0}}

\newcommand{\ke}{\textcolor[rgb]{0.0,0.0,0.0}}

\newcommand{\revised}{\textcolor[rgb]{0.0,0.0,0.0}}

\def\ie{\textit{i.e.}}
\def\eg{\textit{e.g.}}

\def\etal{\textit{et al.}}

\begin{document}

\title{Structure-Informed Shadow Removal Networks}

\author{Yuhao Liu$^{*}$, Qing Guo$^{*\dagger}$, \textit{Member, IEEE}, Lan Fu, Zhanghan Ke, Ke Xu, \\ Wei Feng, \textit{Member, IEEE}, Ivor W. Tsang, \textit{Fellow, IEEE}, Rynson W.H. Lau$^{\dagger}$, \textit{Senior Member, IEEE}
\thanks{This work is supported by two SRG grants from the City University of Hong Kong (Ref:7005674 and 7005843) and one AISG2-GC-2023-008 grant from the National Research Foundation, Singapore, and DSO National Laboratories under the AI Singapore Programme.}
\thanks{Yuhao Liu, Zhanghan Ke, Ke Xu, and Rynson W.H. Lau are with the Department of Computer Science, City University of Hong Kong. (E-mail: yuhaoliu7456@outlook.com; zhanghake2-c@my.cityu.edu.hk; kkangwing@gmail.com;  Rynson.Lau@cityu.edu.hk.)}
\thanks{Qing Guo and Ivor W. Tsang are with the Institute of High Performance Computing (IHPC) and Centre for Frontier AI Research (CFAR), Agency for Science, Technology and Research (A*STAR), Singapore. (E-mail: tsingqguo@ieee.org, ivor\_tsang@cfar.a-star.edu.sg.)}
\thanks{Lan Fu is with InnoPeak Technology Inc. (E-mail: lan.fu@innopeaktech.com.)}
\thanks{Wei Feng is with the College of Intelligence and Computing, Tianjin University, China (E-mail: wfeng@ieee.org.)}
\thanks{$^{*}$Yuhao Liu and Qing Guo are the joint first authors.}
\thanks{$^{\dagger}$Qing Guo and Rynson W.H. Lau are the joint corresponding authors.}}

\markboth{Journal of \LaTeX\ Class Files,~Vol.~14, N.~8, August~2021}%
{Shell \MakeLowercase{\textit{et al.}}: A Sample Article Using IEEEtran.cls for IEEE Journals}


\maketitle

\begin{abstract}
Existing deep learning-based shadow removal methods still produce images with shadow remnants. These shadow remnants typically exist in homogeneous regions with \ryn{low-intensity values}, making them untraceable in the existing image-to-image mapping paradigm.
We observe that shadows mainly degrade images at the image-structure level (in which humans perceive object shapes and continuous colors).
Hence, in this paper, we propose to remove shadows at the image structure level. Based on this idea, we propose a novel structure-informed shadow removal network ({\it StructNet}) to leverage the image-structure information to address the shadow remnant problem.
Specifically, StructNet first reconstructs the structure information of the input image without shadows and then uses the restored shadow-free structure prior to guiding the image-level shadow removal. \ryn{StructNet contains two main novel modules}:
{(1) a {\it mask-guided shadow-free extraction (MSFE) module} to extract image structural features in a non-shadow-to-shadow directional manner, and}
{(2) a {\it multi-scale feature~\&~residual aggregation (MFRA) module} to leverage the shadow-free structure information to regularize \ryn{feature consistency}.}
In addition, we \ryn{also} propose to extend {\it StructNet} to exploit multi-level structure information  ({\it MStructNet}), to further boost the shadow removal performance with minimum computational overheads.
Extensive experiments on three shadow removal benchmarks demonstrate that our method outperforms existing shadow removal methods, \ryn{and our StructNet can be integrated with} existing methods to improve them further.
\end{abstract}


\section{Introduction}
\IEEEPARstart{S}{hadows} exist \ryn{everywhere. They appear on surfaces where light cannot reach due to occlusions}.
\yu{\ryn{Faithfully recovering} the original color and textures of shadow regions \ryn{helps} facilitate many other tasks, \eg, light source analysis~\cite{1265865}, face recognition~\cite{zhang2018improving}, object detection~\cite{Fan_2020_CVPR}, and novel image creation~\cite{ps}.}
Hence, shadow removal is a long-standing problem in computer vision and graphics, with many methods proposed.

Conventional shadow removal methods \ryn{are typically based on modeling} varied intensity~\cite{gryka2015learning} \ryn{and} illumination~\cite{guo2012paired}, \ryn{or involving} user interaction~\cite{gong2016interactive}. 
\yu{They} usually fail when the prior assumptions are not satisfied or the scenes are intricate.

%
%
\begin{figure}[t!]
\centering
\includegraphics[width=9cm, height=4cm]{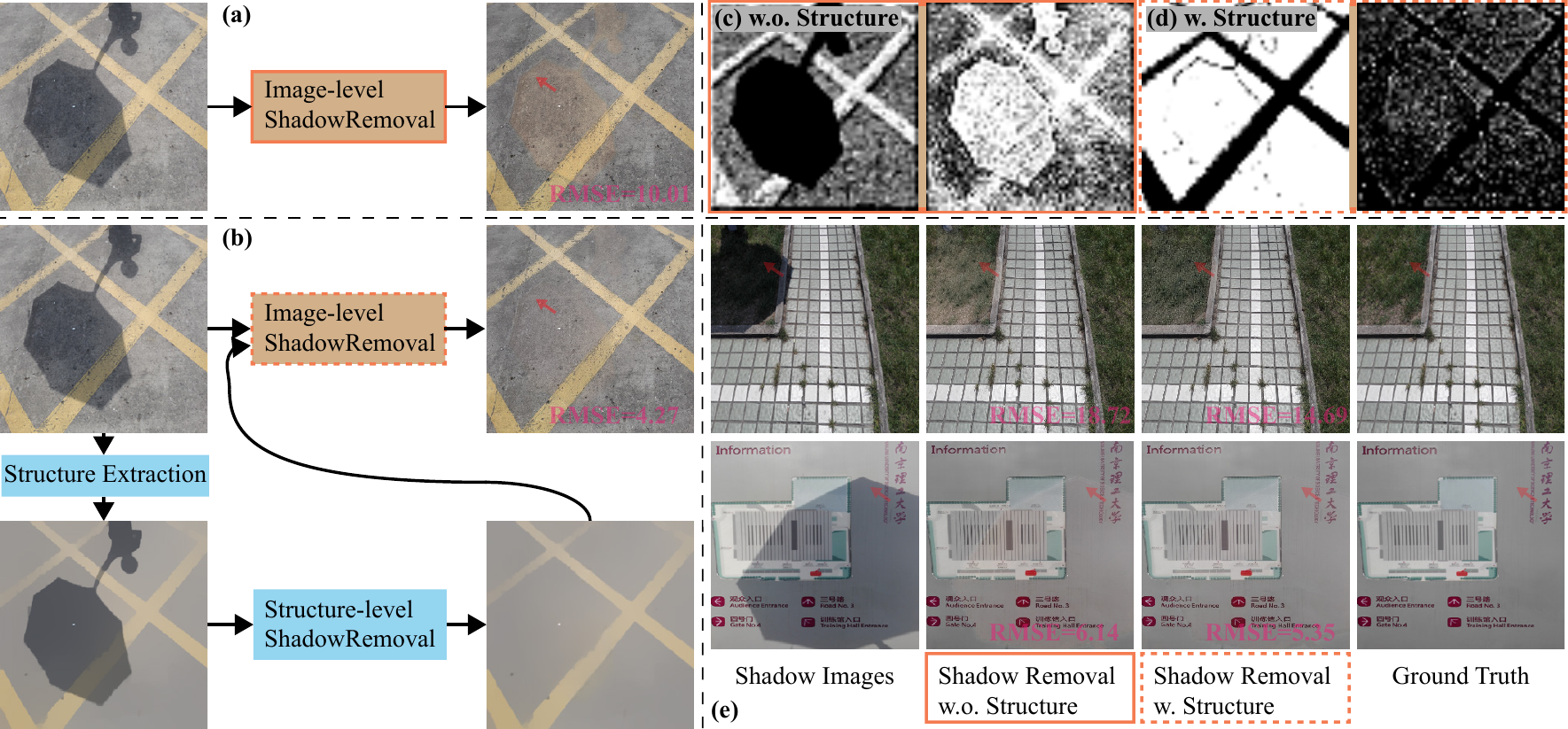}
\caption{(a) \ryn{State-of-the-art shadow removal methods (\eg, AEF~\cite{fu2021auto})} \ryn{typically learn a direct
\yu{shadow-to-shadow-free} mapping and} may often produce shadow remnants with color artifacts. (b) We propose to incorporate image-structure information into the shadow removal process. We visualize the features of \ryn{approaches} (a) and (b) in (c) and (d), respectively, \ryn{which show} that features of \yu{(d)} are structured according to region homogeneity. (e) Results of original AEF and its structure-enhanced counterpart, where red arrows indicate the region \ryn{with} shadow remnants exist, and \yu{RMSE metric are shown for reference.} 
}
\label{fig:motivation}
\vspace{-3mm}
\end{figure}
%
Deep learning-based shadow removal methods \cite{qu2017deshadownet,hu2019direction,wang2018stacked,le2020from,fu2021auto} achieve impressive performances in recent years due to the high generalization capability of advanced neural networks. 
These methods typically formulate the shadow removal  \ryn{problem} as a \yu{shadow-to-shadow-free images mapping.}
Qu \etal \cite{qu2017deshadownet} \yu{first} use CNNs to extract shadow-related information (\ie, location, appearance, and semantic information) \ryn{and then predict} the shadow matte for shadow removal.
Fu \etal \cite{fu2021auto} use CNNs to predict exposure parameters and \ryn{then} remove shadows by fusing multiple shadow exposures. 
However, \ryn{these state-of-the-art methods} may still produce unsatisfactory results \ryn{with shadow remnants and color artifacts}.
In \figref{fig:motivation}(a), we can see yellowish shadow remnants in the result \ryn{from} AEF \cite{fu2021auto}. These remnants are usually internally homogeneous and of low \ryn{intensity values}, making them hard to detect by the existing image-level shadow removal paradigm represented by~\cite{fu2021auto}.

In this work, we propose to address the shadow remnant problem by incorporating the image-structure information \ryn{(which consists of low-frequency image components that represent the object colors and shapes), as shown} in \figref{fig:motivation}(b)).
While the structure \yu{layer} of an image is the primary \ryn{information perceived by} the human vision system~\cite{arnheim1954art,johnson2010our}, \ryn{it separates \yu{the observed} objects into \yu{multiple} homogeneous regions with similar \yu{colors and intensities}~\cite{karacan2013structure}.} 
Hence, \ryn{it should be much easier to locate and much cleaner to remove shadows in the image-structure layer,} due to the absence of high-frequency texture details. \ryn{With the recovered shadow-free image-structure layer as guidance, it may then be possible to restore object details in shadow regions.}

To verify our idea, we \yu{use} the naive UNet~\cite{ronneberger2015U-net} \yu{to first perform} image-structure shadow removal, the output of which is then used to guide the image shadow removal process (\figref{fig:motivation}(b)).
\ryn{With this model, we show} that structure-level shadow removal \ryn{can help boost the performances of a state-of-the-art} shadow removal method~\cite{fu2021auto} (\figref{fig:motivation}\yu{(a) \textit{vs} (b)} and \figref{fig:motivation}(e) column 2 \textit{vs} 3).
We visualize the feature maps of \yu{original AEF in \figref{fig:motivation}(b) and the structure-enhanced counterpart} in \figref{fig:motivation}(c) and \figref{fig:motivation}(d), respectively. 
\ryn{We can see that features in (d) are structured based on} region homogeneity, which \ryn{helps alleviate} the color artifacts of \figref{fig:motivation}(a). 
However, we \ryn{also note} that the standard convolution used in the naive model (as well as in almost all existing methods) adopts spatially-shared weights to process \ryn{both} shadow and non-shadow regions, \ryn{and} neglects their distinct patterns, \ryn{resulting in} color shifts.

Based on the above analysis, we propose the  \textit{structure-informed shadow removal network (StructNet)}, which consists of the structure-level shadow removal step in stage-1 and the image-level shadow removal step in stage-2. 
We propose two novel modules to \yu{facilitate the shadow removal} in the structure-level: {\it mask-guided shadow-free extraction (MSFE)} and {\it multi-scale feature \& residual aggregation (MFRA) modules}. 
\ryn{The MSFE module aims to model non-shadow-to-shadow structure information conditioned on the non-shadow regions, while the MFRA} module focuses on incorporating the extracted shadow-free structure information into the shadow removal process with feature consistency regularization.
They can dynamically extract shadow-free structure information and propagates them into shadow regions for shadow removal.
We conduct extensive experiments on three benchmarks to evaluate  our method and show that StructNet outperforms state-of-the-art \ryn{shadow removal methods}.
\ryn{StructNet can also be incorporated into existing fully-supervised shadow removal methods to help enhance their performances}.
\ryn{Finally, we propose to conduct the shadow removal task at multiple structure levels with a single architecture (named MStructNet), which is not only efficient but also } outperforms state-of-the-art methods.
In summary, we make the following \qing{efforts}:
\begin{itemize}
    \item 
    We construct a naive model (\ie, the vanilla UNet) for structure-level shadow removal and conduct extensive empirical studies on it. We \ryn{show} that removing shadows at the structure level is \ryn{more effective} than that at the image level, and the restored shadow-free structures can \ryn{improve the quality of the output images}.
    \item We propose the \textit{structure-informed shadow removal network (StructNet)}, which contains two novel modules for structure-level shadow removal: mask-guided shadow-free extraction (MSFE) module and multi-scale feature \& residual aggregation (MFRA) module. MSFE learns directional shadow-free structure information  from non-shadow to shadow regions, while MFRA regularizes feature consistency by dynamically fusing the output from MSFE with whole image features.
    \item We further propose a self-contained shadow removal method, multi-level StructNet (MStructNet), which utilizes multi-level shadow structures at the feature level with low parameters for high-quality shadow removal.
    \item Extensive \ryn{evaluations and ablation studies} on three public datasets show that the proposed StructNet can help enhance the performances of existing SOTA methods 
    , and MStructNet achieves high-quality image restoration, outperforming  SOTA shadow removal methods.
\end{itemize}

\section{Related Work}
\label{sec:relatedwork}

\subsection{Shadow Removal}
\label{subsec:relatedwork_shadowremoval}

\textbf{Traditional-based} shadow removal methods \cite{drew2003recovery,finlayson2001,finlayson2005removal,finlayson2002removing,yang2012shadow,zhang2015shadow,xu2017learning} mainly rely on image statistical priors (\eg, gradients and colors).
Finlayson \etal \cite{finlayson2009entropy,finlayson2005removal} solve shadow detection and removal via gradient consistency of illumination invariation. 
Shor and Lischinki \cite{shor2008shadow} propose an illumination-based model in which a pixel-wise relationship between shadow and shadow-free pixel intensities is modeled. 
Guo \etal \cite{guo2012paired} propose a relative illumination model based on paired data modeling. 
However, conventional methods often \yu{fail} when their hand-crafted features do not represent real-world scenes.

\textbf{Deep learning-based} techniques, renowned for their advanced modeling capabilities, have found extensive applications in various vision tasks such as detection \cite{girshick2015fast}, segmentation \cite{liu2021tripartite,liu2023referring}, and generation \cite{guo2021jpgnet,li2022misf}. 
With the availability of large-scale datasets in shadow removal  \cite{qu2017deshadownet,wang2018stacked}, numerous approaches  \cite{wang2018stacked,hu2019direction,cun2020towards,zhang2020cla,zhang2020ris,fu2021auto,zhu2022bijective,wan2022style,li2023leveraging} have been proposed. 
\yu{Typically, these methods model shadow removal as an image-to-image mapping process from shadow image to shadow-free image.} 
DeShadowNet \cite{qu2017deshadownet} first proposes to  use multi-branch CNNs to extract multi-level contexts for shadow removal. 
The follow-ups focused on modeling the shadow formation model~\cite{le2019shadow,le2021physics}, and designing different network architectures and exploiting distinctive properties (\eg, contexts~\cite{hu2019direction}, exposures~\cite{fu2021auto}, residuals~\cite{zhang2020cla}, and illuminations~\cite{zhu2022efficient}).
Unpaired/unsupervised methods \cite{hu2019maskshadowgan,le2020from,liu2021from,jin2021dc,liu2021shadow,he2021unsupervised} have also been proposed to alleviate the labeling cost of paired data \yu{through generative adversarial training and pseudo labels generation}. 
\yu{Nonetheless, these methods may still produce shadow remnants and color artifacts. In this paper, we propose to model the image structure \yu{layer} to handle the shadow remnant problem.}
\revised{Naoto \cite{inoue2020learning} \etal~further propose to generate a synthetic shadow dataset for shadow removal.}

\subsection{Image-structure in Vision Tasks}
\label{subsec:relatedwork_structrestoration}

The \yu{image-structure information} \cite{arnheim1954art,johnson2010our} has been \yu{studied in several vision tasks}. 
Ren \etal \cite{ren2019structureflow} propose to leverage the image-structure information to guide their inpainting method to generate image content in a low-to-high frequency manner. 
Gui \etal \cite{gui2020featureflow} propose to leverage the intermediate image-structure layers to constrain the smoothness of consecutive frames for video interpolation. 
For cartoonization, Wang \etal \cite{wang2020learning} propose to process the image-structure layer separately from the texture layer to maintain harmonious colors.

In this paper, we leverage the image-structure information to locate and track the shadow remnants in homogeneous regions to remove shadows and preserve color consistency.

\begin{figure}[t]
\centering
\includegraphics[width=1.0\linewidth]{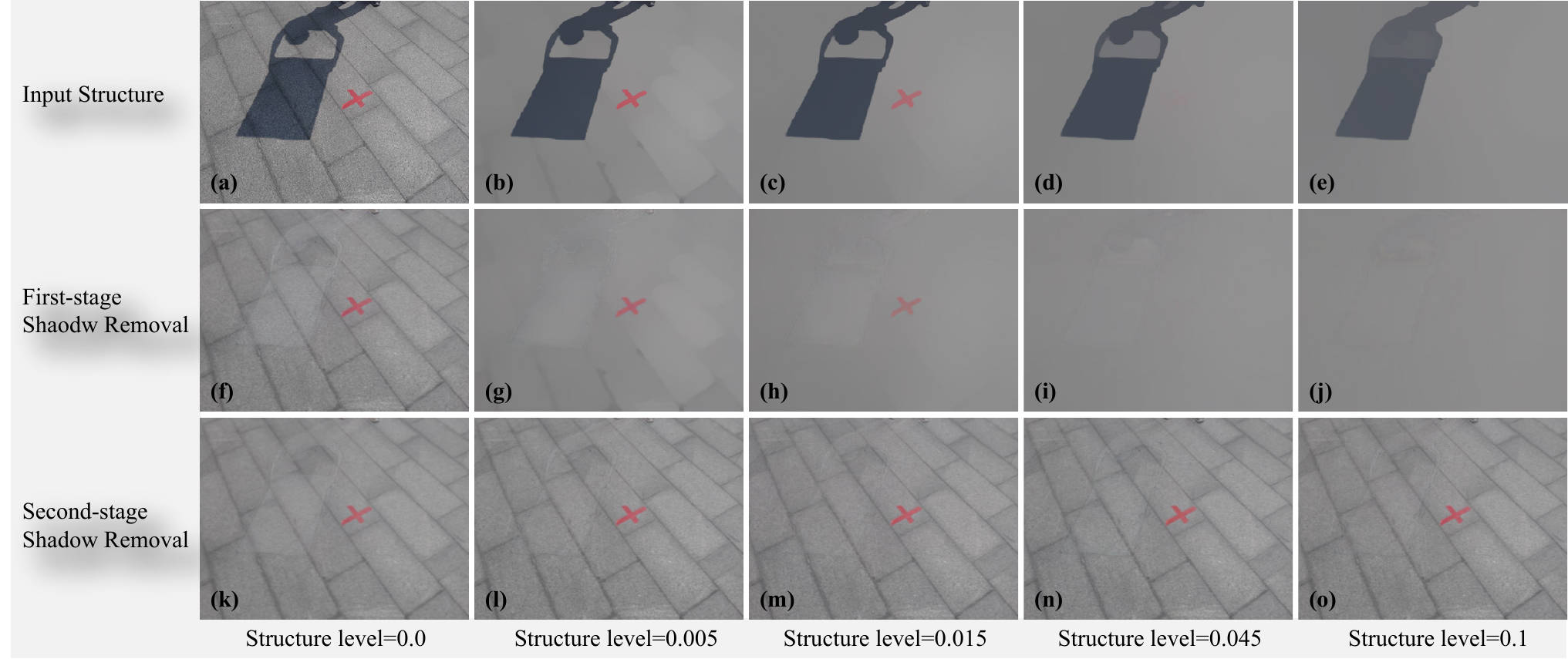}
\caption{
Shadow removal results at different  structure levels. The $1$st row shows the original shadow image (a) and its structures (b)-(e) extracted by~\cite{xu2012structure} at four different structure levels (\ie, $l \in \{0.005, 0.015, 0.045, 0.1\}$). The $2$nd row shows the shadow removal results by feeding the shadow structures in the $1$st row to respective vanilla UNets. Image (f) represents the result of the image-level shadow removal, while images (g)-(j) are the results of structure-level shadow removal with $l>0.0$.
The $3$rd row shows restoration results of \ryn{our naive two-stage shadow removal network} by feeding the restored shadow-free structures (\ie, the images at $2$nd row) into the second vanilla UNets.
}
\label{fig:motcases}
\end{figure}

\section{Structure-level Shadow Removal}
\label{sec:structshadowremoval}

In this section, we \ke{introduce \ryn{our structure-level shadow removal approach}.} \ke{We first} formulate the structure-level shadow removal problem in \secref{subsec:problemformulation} and then investigate the application of structure information in shadow removal in 
 \secref{subsec:empiricalstudy}.

\subsection{\ke{Formulation of Structure-Level Shadow Removal}}
\label{subsec:problemformulation}

In structure-level shadow removal, we first use a  structure extraction method $\varphi(\cdot)$  to map the shadow image $\mathbf{I} \in \mathds{R}^{H \times W \times 3}$ to \ke{\yu{its} structure image/layer, in which \ke{image inherent colors} and main outlines are preserved while detailed textures are removed}
(see \figref{fig:motivation}(b) and \figref{fig:motcases}), as
%
\begin{align} \label{eq:structure}
    \mathbf{S}_l=\varphi(\mathbf{I},l),
\end{align}
%
where $l > 0.0$ is a hyper-parameter determining \ke{the structure level, and $\mathbf{S}_{l}\in\mathds{R}^{H\times W\times 3}$ is the structure image at the $l$th structure level}.
\ke{A higher $l$ will remove} more detailed textures (see the first row of \figref{fig:motcases}).
We follow the setups in \cite{fu2021auto,le2019shadow} to \ke{formulate the $l$th structure-level shadow removal}:
%
\begin{align} \label{eq:struct_removal}
    \hat{\mathbf{S}}_l=\ke{\phi_{l}}(\mathbf{S}_l,\mathbf{M}),
\end{align}
%
where \ke{$\phi_{l}(\cdot)$} is the shadow removal \ke{model corresponding to $\mathbf{S}_l$}, and $\mathbf{M}\in \mathds{R}^{H\times W}$ is a binary mask that indicates shadow and non-shadow pixels with $1$ and $0$, respectively. \ke{Note that the shadow mask is an input to the shadow removal task.}
\ke{The output $\hat{\mathbf{S}}_l$ is \yu{the restored} 
structure \yu{layer} at the $l$th structure level, \ie, the result of structure-level shadow removal.}

\subsection{Empirical Studies}
\label{subsec:empiricalstudy}

\ke{To study how the structure information affects shadow removal results, we employ the structure extraction model proposed by Xu \etal~\cite{xu2012structure} as $\varphi(\cdot)$. We design
a variant of vanilla UNet~\cite{ronneberger2015U-net}, which consists of an encoder with 5 convolution layers and a decoder with 5 de-convolution layers, as $\phi_{l}(\cdot)$.
Each layer in our $\phi_{l}(\cdot)$ is followed by an Instance Norm~\cite{ulyanov2016instance} function and a Leaky-ReLU~\cite{maas2013rectifier} (for the encoder) or ReLU (for the decoder) function. We set the kernel size, padding, and stride of \ke{each layer to} 4, 2, and 1, respectively.} 
\ke{Based on the above network configurations, we aim to answer the following three questions:}
\ding{182} 
how does the capability of shadow removal vary \ke{at} different structure levels? 
\ding{183} whether 
\ke{the structure-level shadow removal results (\ie, \yu{corrected structure}) could guide the image-level shadow removal?}
\ding{184} whether existing \ke{model architectures}
are suitable for structure-level shadow removal?

\begin{figure}[t!]
\centering
\includegraphics[height=3.6cm]{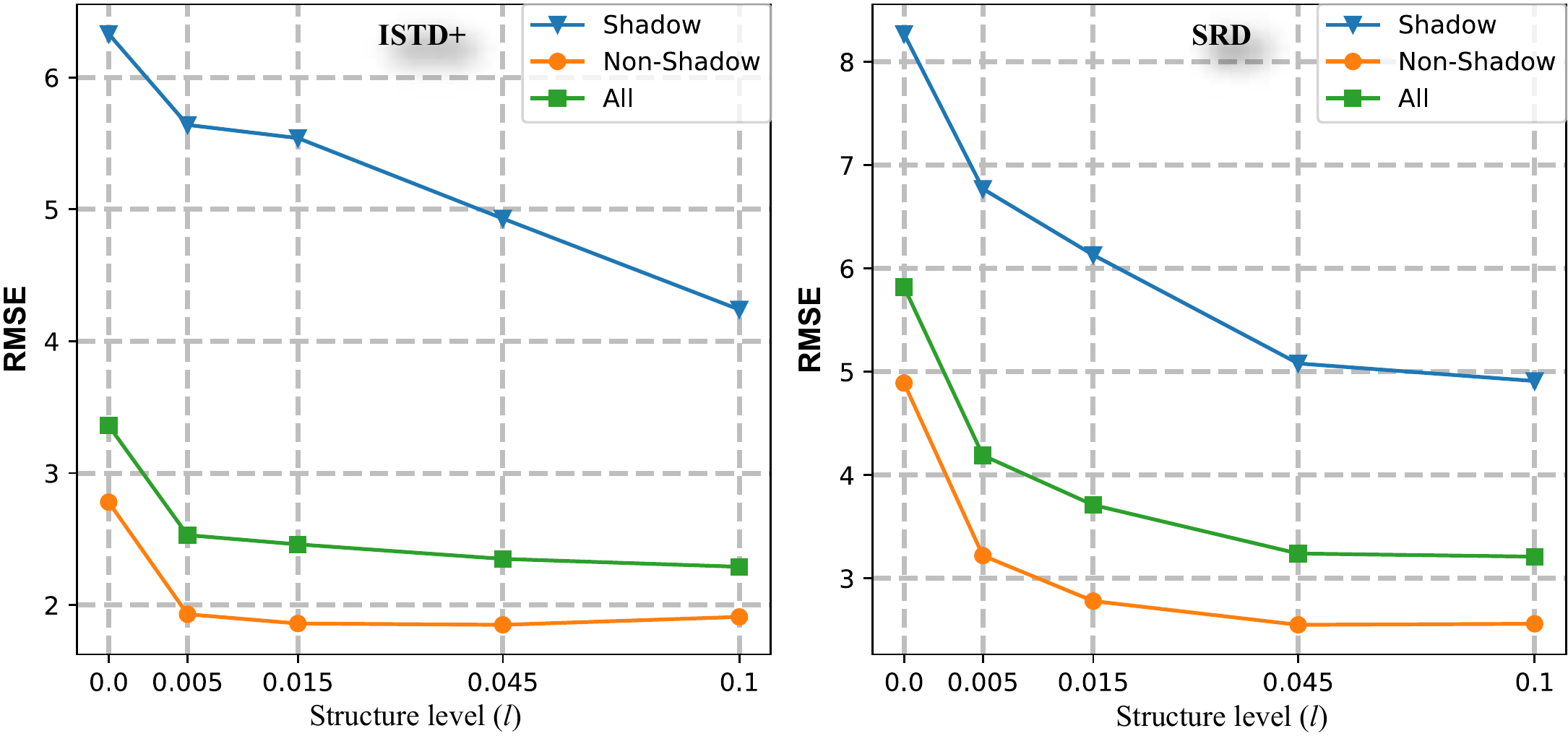}
\caption{Comparison of the image-level (\ie, $l=0.0$) and four structure-level shadow removal process with $l\in\{0.005, 0.015, 0.045,0.1\}$ on two public datasets (\ie, ISTD+ \cite{le2021physics} and SRD \cite{qu2017deshadownet}).
We employ the root mean square error (RMSE) in the LAB color space as the evaluation metric to assess the shadow-removal performances in the shadow regions, non-shadow regions, and the whole (\ie,  All) image, respectively.}
\label{fig:comp_structure_levels}
\end{figure}

\subsubsection{Shadow Removal at Different Structure Levels}
\label{subsubsec:difflevels}

\ke{
Since the shadow removal results may vary at different structure levels ($l$th),
we train and test $\phi_{l}(\cdot)$ at five structure levels $l\in \{0.0, 0.005, 0.015, 0.045, 0.1\}$~\footnote{\yu{The range of structure levels are determined by the structure extraction method. When the level is larger than $0.1$, the image will degrade to a pure color map.}} Note that $l=0.0$ is equivalent to image-level shadow removal, \ie, Eq.\,\ref{eq:structure} with $l=0.0$ as an identity function. 
To avoid the possible influence of elaborately designed loss functions, 
we only optimize the prediction $\hat{\mathbf{S}}_l = \ke{\phi_{l}}(\varphi(\mathbf{I}, l),\mathbf{M})$ via the mean absolute error $L_1(\hat{\mathbf{S}}_l,\mathbf{S}_l^*)=\|\hat{\mathbf{S}}_l-\mathbf{S}_l^*\|_1$, where $\mathbf{S}_l^* = \varphi(\mathbf{I}^*, l)$ is the ground truth structure  generated from the shadow-free image $\mathbf{I}^*$.
On the validation set, we calculate the root mean square error (RMSE) between $\hat{\mathbf{S}}_l$ and $\mathbf{S}_l^*$ after converting them into the LAB color space. The smaller, the better. 
}

\ke{
We conduct evaluations on two widely used datasets, ISTD+ \cite{le2021physics} and SRD \cite{qu2017deshadownet}. Based on the results shown in \figref{fig:comp_structure_levels}, we observe that}
\ding{182} the RMSE on \kk{shadow regions decreases continuously as $l$ increases}. This suggests that it is easier to \ke{obtain high quality shadow removal results} at the structure level (\ie, $l>0$) than at the image level (\ie, $l=0$).
\ke{Such a phenomenon is also reflected in visual results shown in \figref{fig:motcases}, in which there are obvious artifacts in the image-level shadow removal result (\figref{fig:motcases}(f)), but such artifacts are greatly reduced at the structure-level shadow removal results (\figref{fig:motcases}(g)-(j)).}
\ding{183} \kk{The RMSE curves in the non-shadow regions \ryn{descend} at the beginning then become \ryn{flat} when $l$ increases to reach a certain level. The RMSE curves of the whole images have similar shapes to those of non-shadow regions.}
\ke{For non-shadow regions of the ISTD+ dataset, $l=0.1$ has even worse RMSE than that of $l=0.045$ (\figref{fig:comp_structure_levels} left).}

These experiments show that a higher structure level $l$ generally facilitates shadow removal by making the shadow removal network focus more on color and structure information instead of texture information. 
\yu{However, if $l$ is too large, it may lead to shadow spreading, \ie, similar shadow visual patterns may appear in neighboring non-shadow regions (see \figref{fig:motcases}(e)), which in turn causes a higher error in the non-shadow regions.}
%

%
\begin{figure}[t]
\centering
\includegraphics[width=1.0\linewidth]{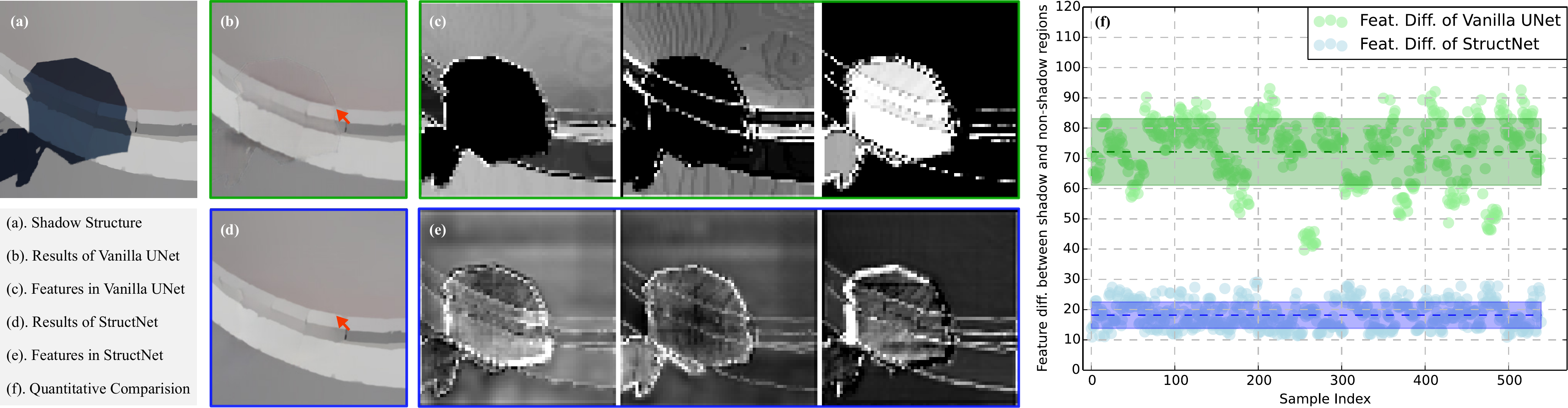}
\caption{
Visualization and quantitative comparison of vanilla UNet and StructNet for structure-level shadow removal. (a) is the input shadow structure \yu{image}, which is fed to the vanilla UNet and StructNet to obtain (b) and (d), respectively. Images (c) and (e) show the randomly sampled three feature channels produced by the $2$nd  convolutional layer of the two networks. In addition, we also extract the features from the $2$nd convolution layer of the vanilla UNet and StructNet of all images in the ISTD+ test set. For each image, we calculate the absolute difference between the shadow and non-shadow regions in each feature channel and obtain the average difference across all channels. Image (f) shows the average feature differences of all images using the vanilla UNet (green points) and StructNet (blue points).}
\label{fig:convdiff}
\end{figure}
%

\begin{table}[t!]
\footnotesize
	\caption{Comparison between direct single-stage image-level shadow removal and five variants ($l=0.0$ is regarded as a special variant.) of two-stage structure-level shadow removal.  
	All experiments are conducted on the ISTD+ dataset with the vanilla UNet and $L_1$ loss function. }
    \centering
     \resizebox{1.0\linewidth}{!}{
            \begin{tabular}{c|c|ccc}
            \toprule
            & \makecell{Structure level $l$ \\ for the first stage} & Shadow~$\downarrow$ & Non-shadow~$\downarrow$ & All~$\downarrow$ \\
            \midrule
            \multirow{5}{*}{\makecell{Two-stage \\ shadow removal}}& 0.0          & 6.28 & 2.99 & 3.53 \\
             \cline{2-5}
            &  0.005    & 5.98 & 2.55 & 3.11 \\  
             & 0.015  & 5.89 & 2.49 & 3.05\\  
             & 0.045    & 6.17 & 2.57 & 3.16 \\ 
             & 0.1      & 6.15 & 2.56 & 3.15 \\ 
            \midrule
             \multicolumn{2}{c|}{\makecell{Direct Single-stage \\  Image-level shadow removal}} & 6.33 & 2.78 & 3.36 \\
            \bottomrule
            \end{tabular}}
	\label{tab:selection_of_structure_levels}
\end{table}

\subsubsection{\ke{Shadow Removal with Structure-level Guidance}}
\label{subsubsec:benefits}
\ke{We investigate if structure-level shadow removal is beneficial to image-level shadow removal and formulate a two-stage pipeline of which
the first stage combines Eq.\,\ref{eq:structure} and \ref{eq:struct_removal} for structure-level shadow removal. The second stage uses a new model $\psi_{l}(\cdot)$, which takes the \yu{corrected structure/layer}  $\hat{\mathbf{S}}_{l}$ as \yu{ancillary input} for image-level shadow removal, as:
\begin{align} \label{eq:image_removal}
    \hat{\mathbf{I}}_{l}=\psi_{l}(\mathbf{I}, \hat{\mathbf{S}}_l, \mathbf{M}),
\end{align}
where $\hat{\mathbf{I}}_{l}$ denotes the image-level shadow removal results guided by $\hat{\mathbf{S}}_l$.
Theoretically, $\psi_{l}$ can be an arbitrary image-level shadow removal method (\eg, ST-CGAN \cite{wang2018stacked} or AEF \cite{fu2021auto}).
\ryn{For simplicity, we simply assume $\psi_{l}$ to have the same architecture as $\phi_{l}$.}
When training the pipeline corresponding to $l\in\{0.0, 0.005, 0.015, 0.045,0.1\}$, we fix $\phi_{l}$ optimized in Sec.\,\ref{subsubsec:difflevels} and learn $\psi_{l}$. We apply the same $L_1$ loss function and RMSE metric as in Sec.\,\ref{subsubsec:difflevels}.
\tableref{tab:selection_of_structure_levels} shows the results (on the ISTD+ dataset) of \ryn{the two-stage shadow removal pipeline} with different $l$ and a single-stage
image-level shadow removal model.
Note that the pipeline with $l = 0.0$ can be regarded as a stack of two vanilla UNet models for image-level shadow removal.} 
\ke{We observe that}
\ding{182} two-stage image-level (\ie, $l=0.0$) shadow removal does not yield better performance, compared to single-stage image-level shadow removal, and shadow remnants cannot be eliminated by simply adding more CNN parameters as shown in \figref{fig:motcases}(a,f,k);
\ding{183} two-stage shadow removal with $l>0.0$ achieves lower RMSE than that of \ke{image-level shadow removal (either two-stage shadow removal with $l=0.0$ or direct single-stage shadow removal)}, which \ke{shows that the restored shadow-free structures can help image-level shadow removal.}
\ke{We also observe from the results in \figref{fig:motcases} that}
the artifacts in \figref{fig:motcases}(f) (\ie, the result of single-stage image-level shadow removal) are \yu{alleviated} by the two-stage shadow removal with $l>0.0$, as shown in \figref{fig:motcases}(l-o)).

\begin{figure*}[t]
\centering
\includegraphics[width=1.0\linewidth]{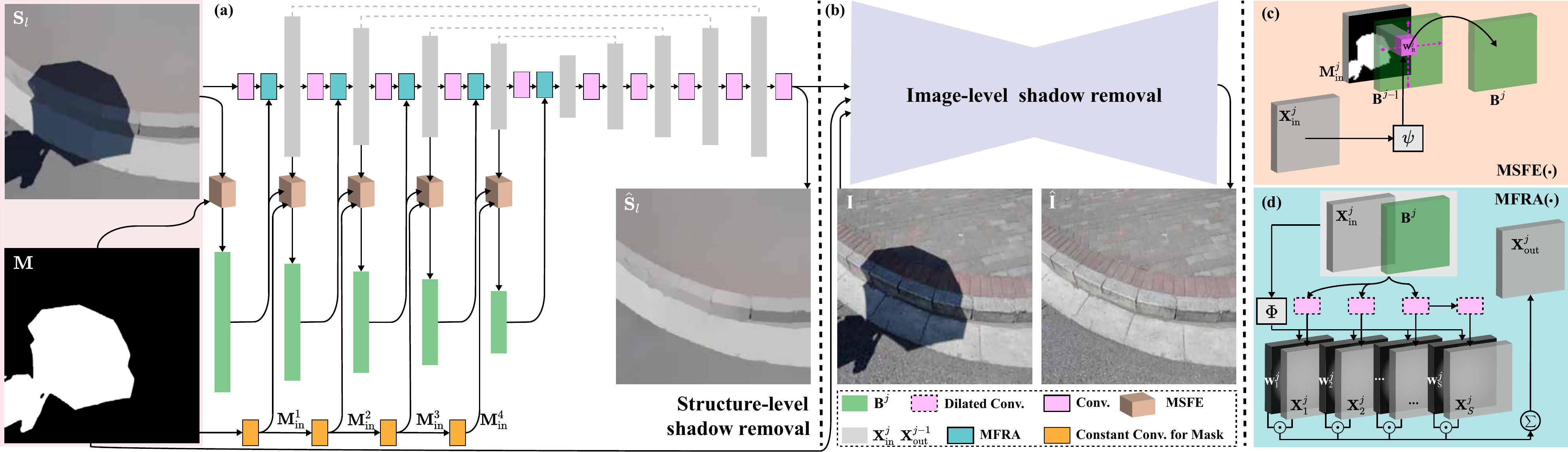}
\caption{Pipeline of the proposed StructNet. (a) shows the structure-level shadow removal. (b) shows the image-level shadow removal with the  assistance of predicted shadow-free structure from (a). (c) and (d) represent the mask-guided shadow-free extraction (MSFE) and the multi-scale feature~\&~residual aggregation (MFRA) modules, respectively, in the architecture.}
\label{fig:structNet_pipeline}
\end{figure*}

\subsubsection{\ke{Limitations of using the Vanilla UNet}}
\label{subsunsection:limi_and_moti}

\ke{
In \secref{subsubsec:benefits}, we have demonstrated that the structure-level shadow removal results can benefit image-level shadow removal to some degree. Here, we would like to know if the vanilla UNet is good enough for this two-stage shadow removal (\ie, first at structure-level and then at image-level).} 
We \ke{observe} that the standard convolution operations used in the vanilla UNet \ke{process} shadow and non-shadow regions uniformly, and ignore the distinctions between them (\eg, color-bias).
\ke{
In other words, the standard convolution used in the vanilla UNet attempts to map shadow and non-shadow regions that have very different appearances  to the same pattern, which makes the learning of the convolution weights challenging.
}
\ke{As a result, the vanilla UNet may produce obvious color shifts between \ke{shadow and non-shadow regions in the output image}, as shown in \figref{fig:convdiff}(b).}

\yu{\ke{To support the above analysis,}}
we visualize three \ryn{randomly} selected feature channels of the $2$nd convolution layer~\footnote{The difference between shadow and non-shadow regions in deeper \ryn{layers} is minimal and indistinguishable. Thus, we choose the 2nd conv.} in the vanilla UNet in \figref{fig:convdiff}(c). We can see that the features of the shadow and non-shadow regions show obvious divergences, although we expect them to be consistent in order to recover the colors of the shadow regions.
We further conduct a quantitative analysis \ke{on the test set of ISTD+}.
\ke{For each sample, we first}
extract the features \ke{output by} the $2$nd convolution \ke{layer}   of the vanilla UNet. 
We then compute the \ryn{means of it feature maps} in the shadow and non-shadow regions separately and show  the absolute difference between the two with a single point in \figref{fig:convdiff}(f).
\ke{We can see that there are huge differences between shadow and non-shadow regions in the feature space.}
Such feature differences are caused by the uniform \ryn{processing} of standard convolutions used in the vanilla UNet. \ryn{As a result, the vanilla UNet produces results with color shift}. 
\ke{This motivates us to design a novel solution to overcome the \ryn{problems} of applying the vanilla UNet to structure-level shadow removal. 
}

\section{StructNet}
\label{sec:StructNet}

In this section, we propose a novel two-stage model, 
named structure-informed shadow removal network (StructNet), to better utilize the structure-level shadow removal results (\ie, \yu{the corrected structure image/layer}) to guide the image-level shadow removal step. 
StructNet contains two novel designs: a mask-guided shadow-free extraction (MSFE) module in \secref{subsec:skipconnect}, and a multi-scale feature $\&$ residual aggregation (MFRA) module in \secref{subsec:msfra}.
The configuration details of StructNet are then described in \secref{subsec:arch_details}.

\revised{
As outlined in \secref{subsunsection:limi_and_moti}, the standard convolution operations in vanilla UNet treat shadow and non-shadow regions uniformly, ignoring the differences between the two regions. Specifically, there is a noticeable feature shifting within the shadow region relative to the non-shadow region.
The standard convolution in vanilla UNet seeks to map shadow and non-shadow regions, which exhibit highly dissimilar appearances, to an identical pattern, which is challenging. For instance, the features from the trained vanilla UNet manifest distinct appearances, as demonstrated in \figref{fig:convdiff} (c).
This discrepancy leads the vanilla UNet to generate evident color shifts between shadow and non-shadow regions in the output image, as depicted in \figref{fig:convdiff} (b).
To rectify the shortcomings of the standard convolution, we propose to make it shadow-aware. We introduce the addition of a directional bridge to the conventional convolution operations, which is guided by the non-shadow regions. This innovative approach promotes homogeneity between the shadow and non-shadow features, thus addressing the issues inherent in the previous method.  
Specifically, given the input features $\mathbf{X}_\text{in}^j\in\mathds{R}^{H_\text{in}^j\times W_\text{in}^j\times C_\text{in}^j}$ at the $j$th layer, we propose to process the features as:
%
\begin{align} \label{eq:structnet_conv}
    \mathbf{X}_\text{out}^j = \text{Fusion}(\mathbf{X}_\text{in}^j*\mathbf{W}^j, \mathbf{B}^j),
\end{align}
%
where $\mathbf{X}_\text{out}^j\in\mathds{R}^{H_\text{out}^j\times W_\text{out}^j\times C_\text{out}^j}$ are the output features, 
$\mathbf{W}^{j}$ are the learnable weights, 
and 
$\mathbf{B}^j\in\mathds{R}^{H_\text{out}^j\times W_\text{out}^j\times C_\text{out}^j}$ is a learned feature shifting tensor aiming to reduce the feature difference between the non-shadow and shadow regions. 
$\text{Fusion}(\cdot)$ is a function to fuse the shifting information in $\mathbf{B}^j$ and the features $\mathbf{X}_\text{in}^j*\mathbf{W}^j$ effectively, thus regularizing the output features to be consistent between the shadow and non-shadow regions.
$\mathbf{B}^j$ is computed by $\text{Bridge}(\cdot)$, as:
%
\begin{align} \label{eq:structnet_conv_res}
\mathbf{B}^j = \text{Bridge}(\mathbf{X}_\text{in}^j, \mathbf{B}^{j-1},\mathbf{M}_\text{in}^j),
\end{align}
%
where $\mathbf{M}_\text{in}^j\in\mathds{R}^{H_\text{in}^j\times W_\text{in}^j}$ is a binary map that indicates the shadow regions with 1's and non-shadow regions with 0's.
$\mathbf{B}^{j-1}$ is the shifting tensor of the previous layer, and $\text{Bridge}(\cdot)$ is trained to extract feature shifting of the non-shadow region shadow regions at the $j$th layer.
Note that such a solution has two benefits: \ding{182} The advantages of the standard convolution are preserved via \reqref{eq:structnet_conv}, which can extract perception across the whole image; \ding{183} The potential shifting between shadow and non-shadow regions is supplemented via \reqref{eq:structnet_conv_res}. 
}

With the above formulation, we propose the structure-informed shadow removal network (StructNet), as shown in \figref{fig:structNet_pipeline}. StructNet consists of two stages. 
\ke{The first stage performs structure-level shadow removal, while the second stage conducts image-level shadow removal guided by the results from the first stage.}
In the first stage, we propose the two novel modules, \ie,  MSFE and MFRA, to extensively exploit the structure information. 
The second stage can be any existing supervised shadow removal method.

\subsection{The MSFE Module}
\label{subsec:skipconnect}

Inspired by the segmentation-aware convolution \cite{harley2017segmentation}, we propose to embed the shadow mask in the convolution operation explicitly and formulate $\text{Bridge}(\cdot)$ as:
%
\begin{align} \label{eq:shadowfreeskip}
\mathbf{B}^j[\mathbf{p}] = \alpha_\mathbf{p}\sum_{\mathbf{q}\in\mathcal{N}_\mathbf{p}}\mathbf{B}^{j-1}[\mathbf{q}](1-\mathbf{M}_\text{in}^j[\mathbf{q}])\mathbf{W}^j_\text{B}[\mathbf{q-p}],
\end{align}
%
where $\mathbf{W}^j_\text{B}\in\mathds{R}^{K^j\times K^j\times C_\text{in}^j \times C_\text{out}^j}$ \ryn{are the weights} of a convolution layer, $\mathbf{p}$ and $\mathbf{q}$ are the coordinates of elements in $\mathbf{X}_\text{in}^j$, $\mathbf{M}^j$, $\mathbf{B}^j$, or $\mathbf{W}^j_\text{B}$.
The set $\mathcal{N}_\mathbf{p}$ contains neighboring elements of $\mathbf{p}$, and its size is equal to the kernel size of $\mathbf{W}_\text{B}^j$ (\ie, $K^2$).
The normalization term $\alpha_\mathbf{p}$ is defined as $\frac{1}{\sum_{\mathbf{q}\in\mathcal{N}_\mathbf{p}}{\mathbf{M}_\text{in}^j[\mathbf{q}]}}$.
The mask (\ie, $\mathbf{M}_\text{in}^j$) is \ryn{obtained} by convoluting the mask from the previous layer (\ie, $\mathbf{M}_\text{in}^{j-1}$) with a constant weight (\ie, $\mathbf{W}_\mathds{1}$ whose elements are one) through $\mathbf{M}_\text{in}^j=\mathbf{M}_\text{in}^{j-1}*\mathbf{W}_\mathds{1}^{j}$.
Intuitively, with Eq.\,\ref{eq:shadowfreeskip}, the output $\mathbf{B}^j$ only rely on the non-shadow regions due to the \yu{restriction} 
of the mask $\mathbf{M}^j$
and can fill the gap across shadow and non-shadow regions.

\begin{figure*}[t]
\centering
\includegraphics[width=1.0\linewidth]{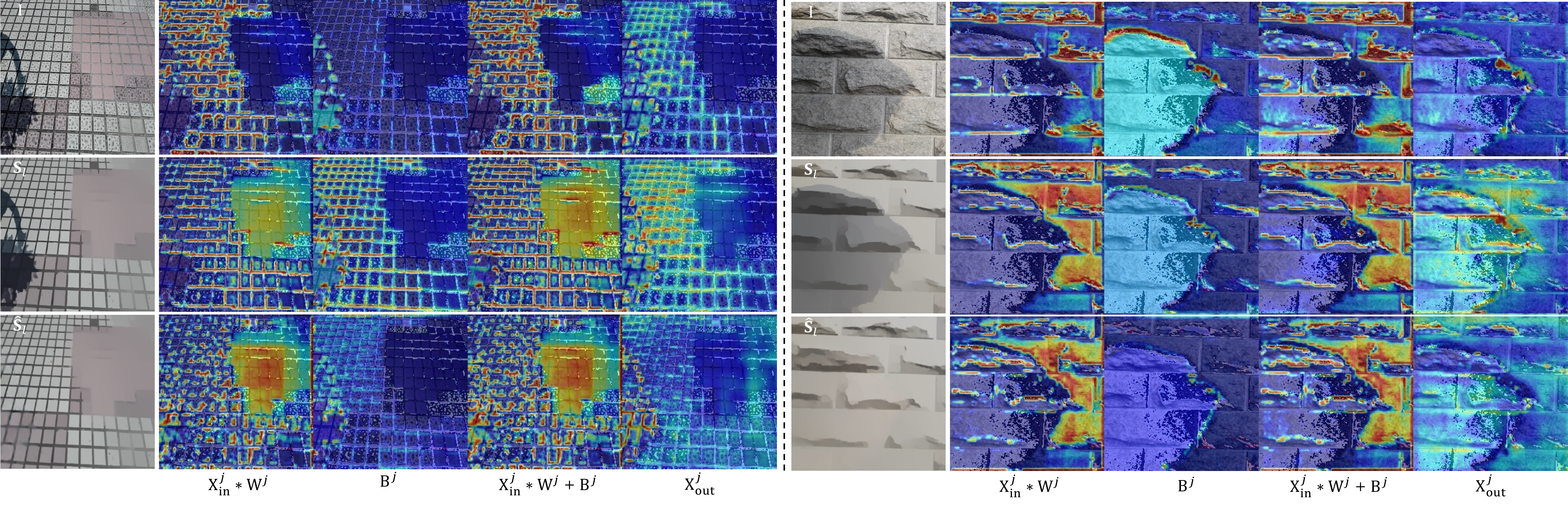}
\caption{\revised{Feature visualization of the global perception ($\mathbf{X}_\text{in}^j*\mathbf{W}^j$), the offset ($\mathbf{B}^j$), direct addition of $\mathbf{X}_\text{in}^j*\mathbf{W}^j$ and $\mathbf{B}^j$,  and output features ($\mathbf{X}_\text{out}^j$) by fusing the former two with MFRA.}}
\label{fig:xwcases}
\end{figure*}

   \revised{ As shown in \figref{fig:xwcases}, 
    due to the low color intensity of the shadow regions,  $\mathbf{X}_{\text{in}}^{j} * \mathbf{W}^{j}$ presents clearer shadow regions and focuses on brighter colors in the non-shadow regions. 
    Instead, $\mathbf{B}^{j}$ can favorably attend to the shadow regions as the shadow mask introduces the positional information. 
    In addition, we also show in column 4 the visualization of the features after adding the standard convolution and shift information (\ie, $\mathbf{X}_{\text{in}}^{j} * \mathbf{W}^{j} + \mathbf{B}^{j}$). The results show that the direct addition of the two fails to achieve feature consistency. This is caused by the fact that each pixel strictly depends on the same position but ignores the convolutional perception and the non-shadow context of the shifting bridge.
    Finally, integrating the global perception $\mathbf{X}_{\text{in}}^{j} * \mathbf{W}^{j}$ and  the offset $\mathbf{B}^{j}$, the result $\mathbf{X}^{j}_{out}$ exhibits consistent homogeneity across the shadow and non-shadow regions.}
In addition, instead of training the weight $\mathbf{W}_\text{B}^j$ for all examples, we propose to make it dynamically \yu{modulated} according to different input features, \ie, $\mathbf{W}^j_\text{B}= \eta(\mathbf{X}_\text{in}^j)$, 
where $\eta(\cdot)$ is a sub-network having two convolution layers.

Our MSFE is different from the partial convolution \cite{liu2018partial} in two aspects: \ding{182} The convolution weights of the proposed bridge function are conditional on the input whole scene features, while those of the partial convolution is fixed after training; \ding{183} The operations with Eq.\,\ref{eq:structnet_conv} and Eq.\,\ref{eq:structnet_conv_res} are a combination of standard and dynamic partial convolution. The former aims to extract the perception of the whole image, while the latter is to bridge the shifting between shadow and non-shadow regions.
%

\subsection{The MFRA Module}
\label{subsec:msfra}

\if
Beyond the element-wise additive fusion of the convolutional perception (\ie, $\mathbf{X}_\text{in}^j*\mathbf{W}^j$) and the shifting $\mathbf{B}^j$, we propose to conduct the fusion at multiple scales with a dynamic aggregation module.
The main motivation stems from the fact that the desired output features should be consistent across shadow and non-shadow regions, and then each desired element in the output features is dependent on the context of the shifting bridge and the convolutional perception around its position.
However, the effective context range for different elements and examples are different, and how to make the fusion adaptive to this change is critical. 
\fi

\revised{
With the extracted convolutional perception (\ie, $\mathbf{X}_\text{in}^j*\mathbf{W}^j$) and the shifting $\mathbf{B}^j$, how to fuse them becomes another critical question. 
Normally, as the network deepens, the shifting features obtained from the non-shadow regions are gradually strengthened, but the extent of this shifting attention is different at different stages in the encoder, \eg, from shallow to deep stages, the shifting gradually spreads from the shadow boundary regions to the whole shadow region.
The naive element-wise additive fusion ignores the feature differences between shadow  and non-shadow regions in  $\mathbf{B}^j$ and $\mathbf{X}_\text{in}^j*\mathbf{W}^j$ at different stages and fails to recover spatial feature homogeneity (see \figref{fig:xwcases} $\mathbf{X}_\text{in}^j*\mathbf{W}^j + \mathbf{B}^j$)). 
Therefore, to allow each stage to account for the features of different scales fully, we adopt a multi-scale fusion strategy.
Specifically, given the convolutional perception (\ie, $\mathbf{X}_\text{in}^j*\mathbf{W}^j$) and the shifting $\mathbf{B}^j$ in Eq.\,\ref{eq:structnet_conv}, we first conduct multiple atrous convolutions with different dilation rates to obtain multi-scale features, \ie,
%
\begin{align} \label{eq:msfra_atrousconv}
\mathbf{X}_{s}^{j} = \sigma([\mathbf{X}_\text{in}^j*\mathbf{W}^j,\mathbf{B}^j]*\mathbf{D}_{s}^{j}),
\end{align}
%
where $\sigma(\cdot)$ is the ReLU function,  $\mathbf{D}_{s}^{j}$ is the weight of an atrous convolution with dilation rate $s$. 
Here, we consider $s\in\mathds{S}=\{1,24,12,6\}$ and get the first three features $\{\mathbf{X}_{s}^{j}|s\in\{1,24,12\}\}$ via Eq.\,\ref{eq:msfra_atrousconv}.
For the last and smallest scale features (\ie, $s=6$), we do not extract from $\mathbf{X}_\text{in}^j$ like Eq.\,\ref{eq:msfra_atrousconv}, but feed $\mathbf{X}_{12}^{j}$ to a dilation convolution to obtain $\mathbf{X}_{6}^{j}$
(see \figref{fig:structNet_pipeline}(d)).
The size of the weights of all $\mathbf{D}_{s}^{j}$ is $3\times 3\times C_\text{in}^j \times C_{\text{out}}^j$ with strides $\{1,1,2,2\}$.
This implementation can alleviate heavy information loss caused by  down-sampling the input features two times directly and reduce the computation cost. 
Then, the key problem is how to combine the four sets of features. 
Since different scales have different regions of interest and different stages require different scale features, we further propose dynamic weight fusion so that different stages can adaptively assign different weights to different scale features. 
To this end, we propose to estimate the combination parameters dynamically according to the inputs, \ie,
%
\begin{align} \label{eq:msfra_comb}
\mathbf{X}_\text{out}^{j} =  \sum_s^S \mathbf{w}_s^j\odot\mathbf{X}_s^j,~\text{with} \nonumber\\
\{\mathbf{w}_s^j|s\in\mathds{S}\} = \Phi([\mathbf{X}_\text{in}^j*\mathbf{W}^j,\mathbf{B}^j]),
\end{align}
%
where $\mathbf{w}_s^j\in\mathds{R}^{H_\text{in}^j\times W_\text{in}^j \times C_\text{in}^j}$ assigns weights for each element in $\mathbf{X}_s^j$. Note that the elements at the same positions but different channels share the same weights. 
$\Phi(\cdot)$ is a subnetwork containing two convolution layers and a softmax layer. A ReLU layer follows each convolution. 
}

\revised{ 
Previous works, such as \cite{cun2020towards}, have explored the fusion of multi-level features for shadow removal. Our multi-scale feature \& residual aggregation (MFRA) approach differentiates itself from \cite{cun2020towards} in the following vital aspects: 
    \ding{182} Diverse Objectives: DHAN aims primarily to learn shadows while retaining low-level details within the input image. This is achieved by aggregating multi-level features through dilated convolutions and the spatial pooling pyramid (SPP). Our MFRA module, however, is crafted to encourage homogeneity between shadow and non-shadow regions within the features.
    \ding{183} Distinct Technical Approaches: DHAN commences by extracting multi-level features and merging them, neglecting to ensure consistency between shadow and non-shadow regions at the feature level. In contrast, our MFRA is embedded within each stage of the encoder, promoting homogeneity at every step by fusing features $\mathbf{X}_{\text{in}}^{j} * \mathbf{W}^{j}$ with feature shifts $\mathbf{B}^{j}$. Additionally, DHAN utilizes SPP coupled with average pooling to aggregate features across different scales, a method that can inadvertently discard essential details. MFRA, on the other hand, operates at each stage without involving pooling operations in the fusion process. It also diverges from DHAN's uniform merging of multi-level features by employing dynamic fusion weights (see \reqref{eq:msfra_comb}) predicted from the input features, allowing for a more adaptable fusion process to various inputs. A detailed comparison of our MFRA module with the SPP used by DHAN is provided in \secref{subsubsec:effective_msfra}.
}

\subsection{Configuration Details}
\label{subsec:arch_details}
In this subsection, we detail the configuration of the first stage of StructNet, which contains three branches. 

The first branch aims to extract the whole scene features and estimate the shadow-free structure prediction. 
It takes the shadow structure (\ie, $\mathbf{S}_l$), the shadow mask (\ie, $\mathbf{M}$), and the shifting predicted by the second branch as inputs.
It consists of one encoder, one decoder, and the proposed MFRA module, where the first two parts  share the same settings with the vanilla UNet in \secref{subsec:empiricalstudy}.
In terms of the fusion function (\ie, MFRA), we set the kernel size  of all convolutional layers to 3, and the number of  kernels/filters (\ie, $C_{out}^{j}$) is $\{64, 128, 256, 512, 512\}$ except for the second layer of  weight generation (\ie,~$\Phi(\cdot)$) where the number of  kernels is equal to the number of parallel branches (\ie, 4).

\yu{The second branch  is to estimate the shifting (\ie, $\mathbf{B}^j$). 
It takes the shifting from the previous layer (\eg, $\mathbf{B}^{j-1}$), the shadow mask from the third branch (\ie, $\mathbf{M}^j_\text{in}$), and the $j$th features from the first branch (\ie, $\mathbf{X}^j$) as inputs. 
It includes five convolutional layers, each corresponding to one of the encoder layers in the first branch and having the same strides and feature dimensions. 
The kernel size $K^{j}$ of each  layer  is $\{7,5,3,3,3\}$, and each layer is followed by a Batch-Norm  and a ReLU function. 
For the $\eta(\cdot)$, we set the kernel size and stride of the two  layers as $\{K^{j},1\}$ and $\{2,1\}$, respectively.}

The third branch takes the shadow mask $\mathbf{M}$ as input and generates distinct binary masks for each layer along the encoder. 
We fix the constant convolutional kernels $\mathbf{W}_\mathds{1}^{j}$ of size $K^{j}$ and with stride 2.

In addition, as shown in Table \ref{tab:selection_of_structure_levels}, the maximum performance gain  is delivered when the structure level is 0.015. 
 Thus, unless otherwise stated, we set $l=$ 0.015 in the proposed StructNet.

\section{Multi-level StructNets (MStructNet)}
\label{sec:MStructNet}

\begin{figure}[t]
\centering
\includegraphics[width=8cm, height=5cm]{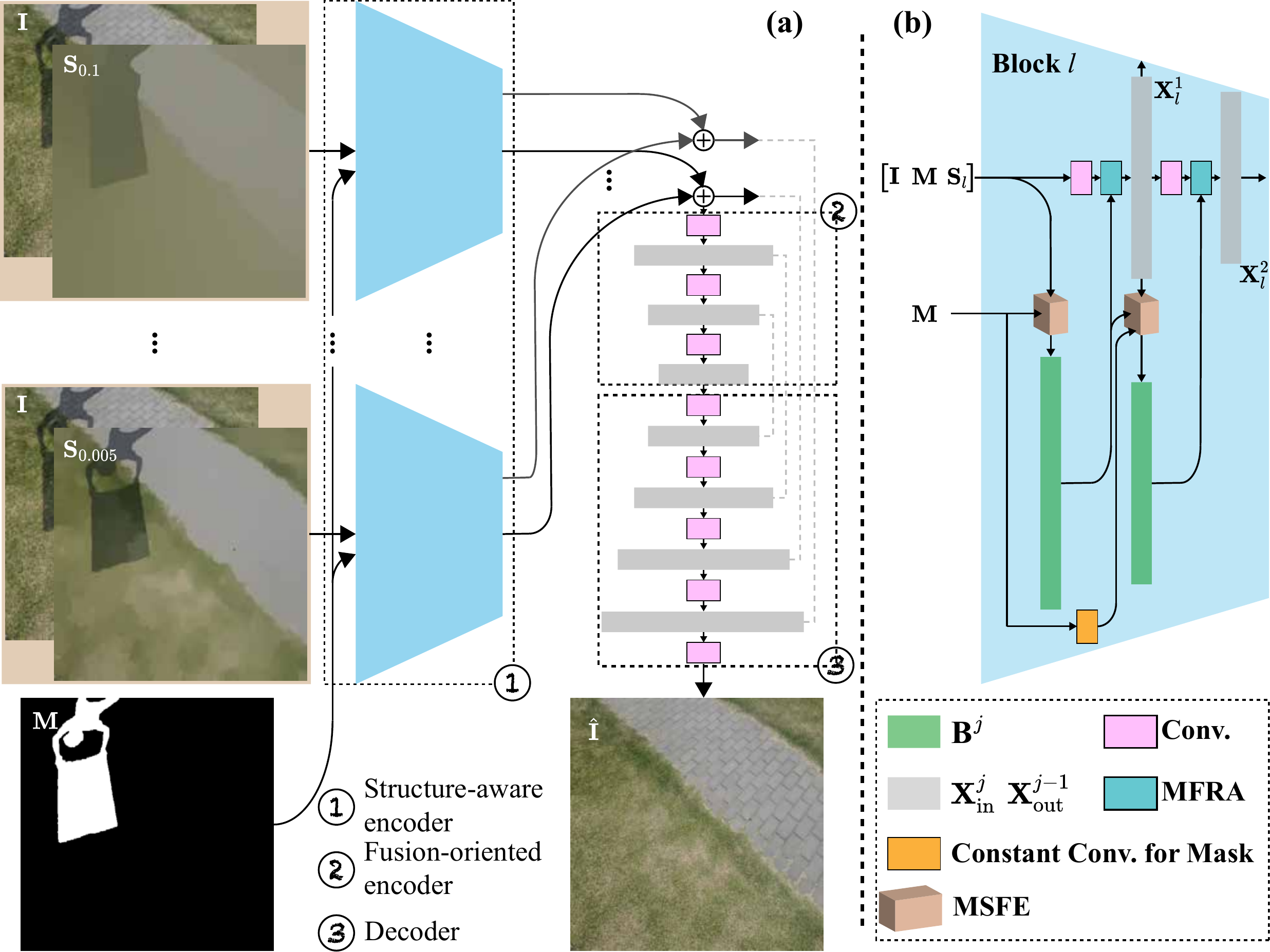}
\caption{Pipeline of the multi-level StructNet (MStructNet). (a) presents the whole pipeline, while (b) shows the detail of the blue blocks in (a).}
\label{fig:mstructnet}
\end{figure}

\ryn{Although our StructNet presented in \secref{sec:StructNet} is able to restore the shadow structure effectively and performs better than the vanilla UNet, benefiting the image-level shadow removal step significantly, such a two-stage solution leads to large computational overheads due to the naive combination of two networks.}
To address this problem, we further propose a self-contained shadow removal method that utilizes multi-level structures at the feature level with \ryn{only a small increase in the parameter numbers}. 
Specifically, we omit the step for predicting the shadow-free structure image through the first stage of StructNet but use the non-shadow structure information directly.
We refer to this method as MStructNet, and show  the pipeline in \figref{fig:mstructnet}.

\subsection{Pipeline}
\label{subsec:mstructnet_pipeline}

Given a shadow image $\mathbf{I}$, we extract structures via Eq.\,\ref{eq:structure} and consider four levels, \ie, $l\in\mathcal{L}=\{0.005, 0.015, 0.045,0.1\}$, to obtain four levels of structure, $\{\mathbf{S}_l|l\in\mathcal{L}\}$.
MStructNet takes the original shadow image,  shadow mask, and  all levels of structure \yu{images/layers} as inputs to predict the shadow-free image directly. 
The whole pipeline contains three components, \ie, structure-aware encoder, fusion-oriented encoder, and decoder. 
The structure-aware encoder contains $|\mathcal{L}|$ blocks to address $|\mathcal{L}|$ structures.
Each block follows the design in StructNet  to ensure that the shadow elements  harmonize with the shadow-free elements. 
The fusion-oriented encoder consists of standard convolutions and is to further extract deep feature embedding \ryn{from the} structure-aware features.
The decoder is to map the feature embedding to the shadow-free image.
We show the whole pipeline in \figref{fig:mstructnet}(a). 
\ryn{As the main difference between this pipeline and the vanilla UNet lies in the structure-aware encoder as shown in  \figref{fig:mstructnet}(b), we discuss the design of this encoder in detail below.}

In terms of the $l$th block in the structure-aware encoder, we have the original shadow image $\mathbf{I}$, the $l$th structure $\mathbf{S}_l$, and shadow mask $\mathbf{M}$ as inputs. 
We feed them to the block having two convolutional layers equipped with the proposed MSFE and MFRA modules (see \figref{fig:mstructnet}(b)), which produce two features denoted as $\mathbf{X}_l^1$ and $\mathbf{X}_l^2$ corresponding to the outputs of the first and second convolution layers, respectively.
For the four levels of structures (\ie, $\{\mathbf{S}_l|l\in\mathcal{L}\}$), we obtain eight output features, \ie, $\{\mathbf{X}_l^1\}_{l\in\mathcal{L}}$ and $\{\mathbf{X}_l^2\}_{l\in\mathcal{L}}$.
We combine the four sets of features in $\{\mathbf{X}_l^1\}_{l\in\mathcal{L}}$ or $\{\mathbf{X}_l^2\}_{l\in\mathcal{L}}$ via an  addition. The combined features are fed to the fusion-oriented encoder and decoder to estimate the shadow-free image $\hat{\mathbf{I}}$.

\subsection{Configuration Details}
\label{subsec:mstructnet_arch}

Same as the first stage of StructNet in \secref{sec:StructNet}, each block \textit{l} in the structure-aware encoder in MStructNet also has three branches.
Take the $l$th block as an example, the inputs to the first branch include shadow image $\mathbf{I}$, shadow structure $\mathbf{S}_{l}$ with level \textit{l} and shadow mask $\mathbf{M}$. 
The three inputs are concatenated along the channel axis and further fed to the standard convolution to perceive the global scene.
The inputs to the second branch are shadow structure $\mathbf{S}_{l}$ and shadow mask $\mathbf{M}$. Then, with the global perceptual features of the first branch as the guiding weights, the shifting features can be obtained by Eq.\,\ref{eq:shadowfreeskip}.
The third branch updates the shadow mask $\mathbf{M}^{j}$ in the same way as in \secref{subsec:arch_details}.
Regarding the fusion-oriented encoder and the decoder, they contain only standard convolutional layers, instance-norm and activation function (\eg, Leaky-ReLU or ReLU), and all the settings are the same as those in the vanilla UNet in \secref{subsec:empiricalstudy}.

\section{Experiment}
\label{sec:experiment}

\subsection{Loss Functions}
\label{subsec:loss}

We train StructNet and MStructNet using $L_1$  and the perceptual loss.
Given a restored image $\hat{\mathbf{I}}$ and its ground truth $\mathbf{I}^*$, we have:  
\begin{align} \label{eq:loss_overall}
    L(\hat{\mathbf{I}},\mathbf{I}^*) = \lambda_{1}L_1(\hat{\mathbf{I}},\mathbf{I}^*)  + \lambda_{2}L_\text{perc}(\hat{\mathbf{I}},\mathbf{I}^*),
\end{align}
where the $\lambda_{1}$ and $\lambda_{2}$ are the coefficients. 
\revised{The $\mathit{L}_{1}(\hat{\mathbf{I}}, \mathbf{I}^{*})$ is the primary loss item to supervise the training process, and we set the $\lambda_{1}=1$ by default. Regarding the $L_\text{perc}(\hat{\mathbf{I}},\mathbf{I}^*)$ loss, we follow Zhang \etal \cite{zhang2018perceptual} to set the $\lambda_{2}=0.1$. }
$L_1(\hat{\mathbf{I}},\mathbf{I}^*)$ is the $L_1$-norm distance to ensure pixel-level visual consistency.  $\mathcal{L}_\text{perc}$  is the perceptual loss \cite{johnson2016perceptual}, which aims to ensure the restored image has the same perception as the ground truth: 
\begin{align} \label{eq:loss_perceptual}
L_{\text {perc}}(\hat{\mathbf{I}}, \mathbf{I}^*)=\sum_{i=1}^3\|\text{VGG16}_{i}(\hat{\mathbf{I}})-\text{VGG16}_{i}(\mathbf{I}^{*})\|_{1},
\end{align}
where $\text{VGG16}_{i}(\cdot)$ represents the activation map of the $\textit{i}$th max-pooling layer in the VGG16 \cite{simonyan2014very} pretrained on ImageNet \cite{krizhevsky2012imagenet}. 

We employ $L(\hat{\mathbf{I}},\mathbf{I}^*)$ to end-to-end train MStructNet directly.
For StructNet, we use the same loss but with $<\hat{\mathbf{S}}_l,\mathbf{S}^*_l>$ to train the first stage, \ie, $L(\hat{\mathbf{S}}_l,\mathbf{S}^*_l)$.
After that, we fix the parameters of the first-stage network and use $L(\hat{\mathbf{I}},\mathbf{I}^*)$ to train the second-stage network.

\subsection{Datasets and Metrics} 
\label{subsec:dataset_metrics}
\textbf{Datasets.} 
We conduct our experiments on three shadow removal benchmark datasets, \ie, SRD \cite{qu2017deshadownet}, ISTD \cite{wang2018stacked} and ISTD+ \cite{le2021physics}, to evaluate the effectiveness of the proposed methods. 
SRD \cite{qu2017deshadownet} is the first large-scale shadow removal dataset, consisting of 3,088 paired shadow and shadow-free images, of which 2,680 are for training and 408 for testing.
Since shadow masks are not available in SRD, we follow AEF \cite{fu2021auto} to utilize Otsu's algorithm to extract the shadow masks from the difference between the shadow and shadow-free images. 
We adopt the extracted masks for  training and testing and use the available masks from DHAN \cite{cun2020towards} for metric evaluation.
The ISTD dataset \cite{wang2018stacked} contains 1,870 triplets (\ie, shadow image, shadow mask, and shadow-free image) for shadow removal, with 1,330 for training and 540 for testing.
Le \etal  \cite{le2019shadow} later corrected the color consistency in ISTD  to form the ISTD+ dataset. 
For both ISTD and ISTD+, we follow AEF \cite{fu2021auto} to use the ground-truth shadow masks for training and extracted masks from Otsu's algorithm for testing.

\textbf{Evaluation Metrics.} 
\revised{We follow methods \cite{hu2019direction,fu2021auto} to compute the root mean square error (RMSE) between the shadow-removed image and ground-truth shadow-free image in the LAB color space, which is also named image-level RMSE.
When evaluating structure-level shadow removal, we compute RMSE between the predicted and ground truth structures as described in \secref{subsec:empiricalstudy}, which is denoted as structure-level RMSE.
We also report the peak signal-to-noise ratio (PSNR) and structural similarity index (SSIM). 
In addition, we also adopt Learned Perceptual Image Patch Similarity (LPIPS)~\cite{zhang2018perceptual} to evaluate the perceptual quality of the shadow-free prediction. The lower the LPIPS, the higher the perceived quality. 
Note that all metrics are computed in the shadow region (S.), non-shadow regions (N.~S.), and the whole image (All), respectively.}

\subsection{Method Settings} 
\label{subsec:setups}

\textbf{Baseline methods enhanced by StructNet.} Our StructNet proposed in \secref{sec:StructNet} is able to enhance existing shadow removal methods by first conducting structure-level shadow removal, and then using the restored shadow-free structure as an auxiliary prior \ryn{for the baseline method} to predict the shadow-free image in the second stage. 
\ryn{We regard four baseline methods (\ie, the vanilla UNet in \secref{sec:structshadowremoval}, STCGAN \cite{wang2018stacked}, AEF \cite{fu2021auto} and SADC \cite{xu2022shadow}) as the second-stage networks in StructNet, resulting in four variants.}
We chose these methods due to their distinct frameworks, highlighting StructNet's exceptional extensibility and flexibility. 
Note that the four versions share the same structure-level shadow removal network. We fix the first-stage network and retrain \ryn{only} the second-stage networks.

To further validate the advantages of the structure-informed shadow removal networks, we compare the StructNet variants and MStructNet with 
two traditional methods: Guo \etal \cite{guo2012paired}, Gong \etal \cite{gong2016interactive}, and eighteen deep learning-based methods: DeshadwoNet \cite{qu2017deshadownet}, STCGAN \cite{wang2018stacked}, DSC \cite{hu2019direction}, MS-GAN \cite{hu2019maskshadowgan}, AR-GAN \cite{ding2019argan}, SP+M-Net \cite{le2019shadow}, CLA \cite{zhang2020cla}, RIS \cite{zhang2020ris}, Param+M+D-Net \cite{le2020from},  DHAN \cite{cun2020towards}, G2R \cite{liu2021from}, AEF \cite{fu2021auto}, DC-GAN \cite{jin2021dc}, SP+M+I-Net \cite{le2021physics}, BMNet \cite{zhu2022bijective}, SADC \cite{xu2022shadow}, EMD-Net \cite{zhu2022efficient}, and SGNet \cite{wan2022style}.

\begin{table*}[!htbp]
	\caption{Validation results of StructNet-equipped shadow removal methods on ISTD and ISTD+ datasets. We embed four existing models, \ie, vanilla UNet, STCGAN \cite{wang2018stacked}, AEF \cite{fu2021auto} and SADC \cite{xu2022shadow}, in our StructNet framework as four variants, and compare them with the original methods.
	}
	\begin{center}
		\renewcommand\tabcolsep{8pt}
		\renewcommand\arraystretch{1.}
            \begin{tabular}{l|l|ccc|ccc|ccc}
            \toprule
        \multirow{2}{*}{Datasets} & \multirow{2}{*}{Methods} & \multicolumn{3}{c|}{RMSE~$\downarrow$} & \multicolumn{3}{c|}{PSNR~$\uparrow$} & \multicolumn{3}{c}{SSIM~$\uparrow$}   \\
             & & S. & N.S. & All & S. & N.S. & All & S. & N.S. & All \\
            \midrule
            \multirow{8}{*}{ISTD+ \cite{le2019shadow}} &
            vanilla UNet & 5.89 & 2.49 & 3.05 & 37.86 & 38.15 & 34.34 & 0.990 & 0.985 & 0.970	 \\
            & StructNet-UNet &5.31 &2.52 & 2.97 & 38.43 & 37.33 & 34.31 & 0.990 & 0.979 & 0.963 	 \\ 
            \cline{2-11}
            & STCGAN~\cite{wang2018stacked} &9.39	&4.25 &5.09 &35.09	&33.92	&30.36	&0.983	&0.961	&0.937	 \\
             & StructNet-STCGAN &6.25	&3.58	&4.02 &37.44	&34.52	&32.00	&0.988	&0.968	&0.949	 \\ 
             \cline{2-11}
             & AEF\cite{fu2021auto} & 6.55 & 3.77 & 4.23 & 36.04 & 31.16 & 29.44 & 0.978 & 0.892 & 0.861   \\
             & StructNet-AEF &6.35 &3.75 & 4.17 & 36.08 &31.18 &29.52 &0.978 &0.892 &0.861  \\ 
             \cline{2-11}
             & SADC \cite{xu2022shadow} &6.21	&3.05	&3.57 & 37.18	&37.69	&33.88	&0.991	&0.982	&0.968	 \\
             & StructNet-SADC &5.82	&2.83	&3.32 & 37.92	&37.72	&34.26	&0.991	&0.983	&0.969	 \\                   
	       \midrule                        
\multirow{8}{*}{ISTD \cite{wang2018stacked}} & 
	   vanilla UNet & 7.29 & 4.73 & 5.09 & 35.69 & 31.70 & 29.74 & 0.987 & 0.970 &	0.951  \\
            & StructNet-UNet & 6.33 &4.71 &4.98 & 36.60 & 31.57 & 29.94 & 0.988 &0.970 &0.952	 \\ 
             \cline{2-11}
             & STCGAN \cite{wang2018stacked}  &10.11 &5.76  &6.47 &33.93 &30.18 &27.90 &0.981 &0.959 &0.932  \\ 
             & StructNet-STCGAN  &7.52 &5.64  &5.95 & 35.46 &30.52 &28.75 &0.985 &0.961 &0.939  \\ \cline{2-11}
             & AEF \cite{fu2021auto}   & 7.98 & 5.54  & 5.94 &34.39  & 28.61  & 27.11 &0.974  &0.880  & 0.844  \\ 
             & StructNet-AEF  &7.49 &5.67  &5.97 & 34.72 &28.09 &26.86 &0.975 &0.880 &0.844  \\ \cline{2-11}
             & SADC \cite{xu2022shadow}   &7.19	&5.06	&5.41 & 35.52 & 31.97 & 29.85	&0.989	&0.976	&0.961	 \\ 
             & StructNet-SADC &6.83	&4.69 &5.04 & 36.40	&32.27	&30.32	&0.989	&0.978	&0.963	 \\
            \bottomrule
            \end{tabular}
    \end{center}
	\label{tab:structnet_results}
\end{table*}

\vspace{-3mm}
\subsection{Comparisons to the state-of-the-arts.}
\label{subsec:comparison_sotas}
%

\begin{table}[t]
	\caption{Quantitative comparison with the SOTA methods on the ISTD dataset.  `-' indicates values that are not available. 
	The best results are highlighted in bold.}
	\begin{center}
		\renewcommand\tabcolsep{1.0pt}
		\renewcommand\arraystretch{1.}
        \begin{tabular}{l|ccc|ccc|ccc|c}
        \toprule
         \multirow{2}{*}{Methods}  &
         \multicolumn{3}{c|}{RMSE~$\downarrow$} & \multicolumn{3}{c|}{PSNR~$\uparrow$} & \multicolumn{3}{c}{SSIM~$\uparrow$} & \multirow{2}{*}{\revised{LPIPS $\downarrow$}}  \\
         & S. & N.S. & All & S. & N.S. & All & S. & N.S. & All \\
         \midrule
         Guo \textit{et al.}~\cite{guo2012paired} & 18.65 & 7.76 & 9.26 &27.76 & 26.44 & 23.08 & 0.964 & 0.975 &0.919 &-  \\ 
        STCGAN~\cite{wang2018stacked}  &10.11 &5.76  &6.47 &33.93 &30.18 &27.90 &0.981 &0.959 &0.932 &0.092\\  
          
         MS-GAN~\cite{hu2019maskshadowgan} &10.57 & 5.91 & 6.67 &31.73 &29.02 & 26.36 &0.980 &0.959 &0.928 &-  \\ 
          DSC~\cite{hu2019direction} & 8.45 & 5.03 & 5.59  &34.64 & 31.26 & 29.00 & 0.984 & 0.969 & 0.944 & 0.223\\ 
          DHAN \cite{cun2020towards} & 7.49 & 5.30 & 5.66  &35.53 &31.05 & 29.11& 0.988 & 0.971 & 0.954 & 0.089  \\ 
          AR-GAN \cite{ding2019argan}&7.21&5.83&6.68&-&-&-&-&-&-&-\\
          RIS \cite{zhang2020ris} &8.99  & 6.33 & 6.95 &-&-&-&-&-&-&-\\
          CLA \cite{zhang2020cla} &9.01  & 6.25 & 6.62 &-&-&-&-&-&-&-\\
          CANet \cite{chen2021canet} & 8.86 & 6.07 & 6.15 &-&-&-&-&-&-&- \\
          DC-GAN~\cite{jin2021dc} & 10.55 & 5.79 & 6.57  &31.69 & 28.99 & 26.38 & 0.976 & 0.958 & 0.922 &0.121\\ 
          BMNet \cite{zhu2022bijective} & 7.60 & 4.59 & 5.02 &35.61 & \textbf{32.80} & 30.28 & 0.988 & \textbf{0.976} &\textbf{ 0.959} & 0.089 \\
           EMNet \cite{zhu2022efficient}& 8.08 & 4.75 & 5.22 & 36.27 & 31.85&  29.98 & 0.986 & 0.965 & 0.944 &0.087 \\
          \midrule
          StructNet & \textbf{6.33} &4.71 & 4.98 & 36.60 & 31.57 & 29.94 & 0.988 &0.970 &0.952 &0.072\\ 
          MStructNet  & 6.34 & \topone{4.35} & \topone{4.68} &\toptwo{36.85} & 32.49 & \topone{30.65} & \toptwo{0.989} & 0.972 & 0.955 &\textbf{0.059}\\ 

            \bottomrule
            \end{tabular}
    \end{center}
	\label{tab:istd_comp_sotas}
\end{table}

\subsubsection{Validation Results}
\label{subsubsec:structnet_valid}

 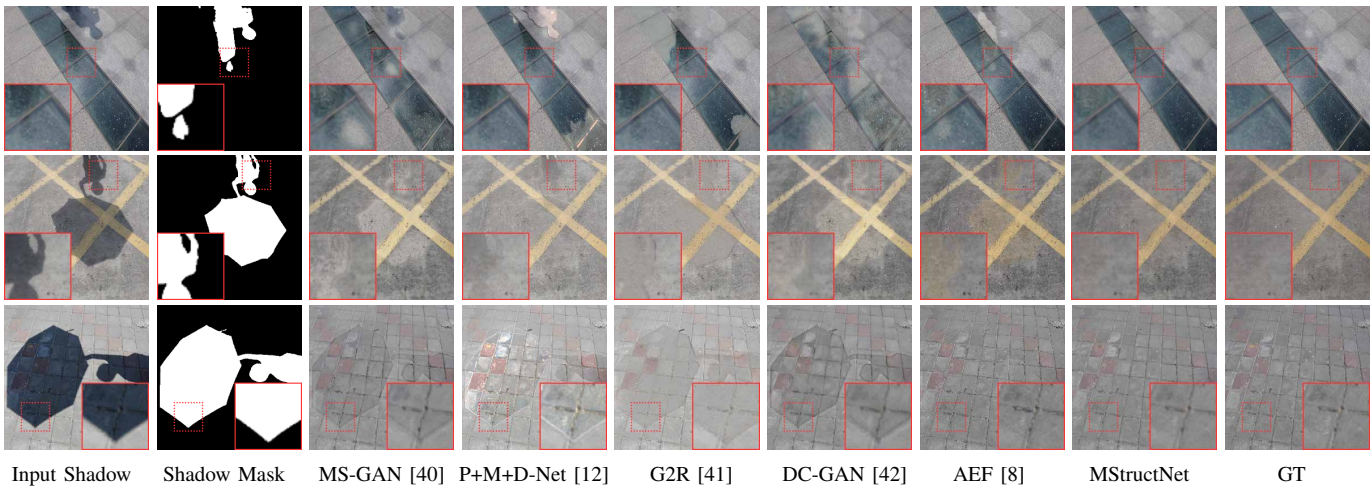
\begin{figure*}[!th]
 \centering
 \begin{overpic}
    [width=1.0\linewidth]{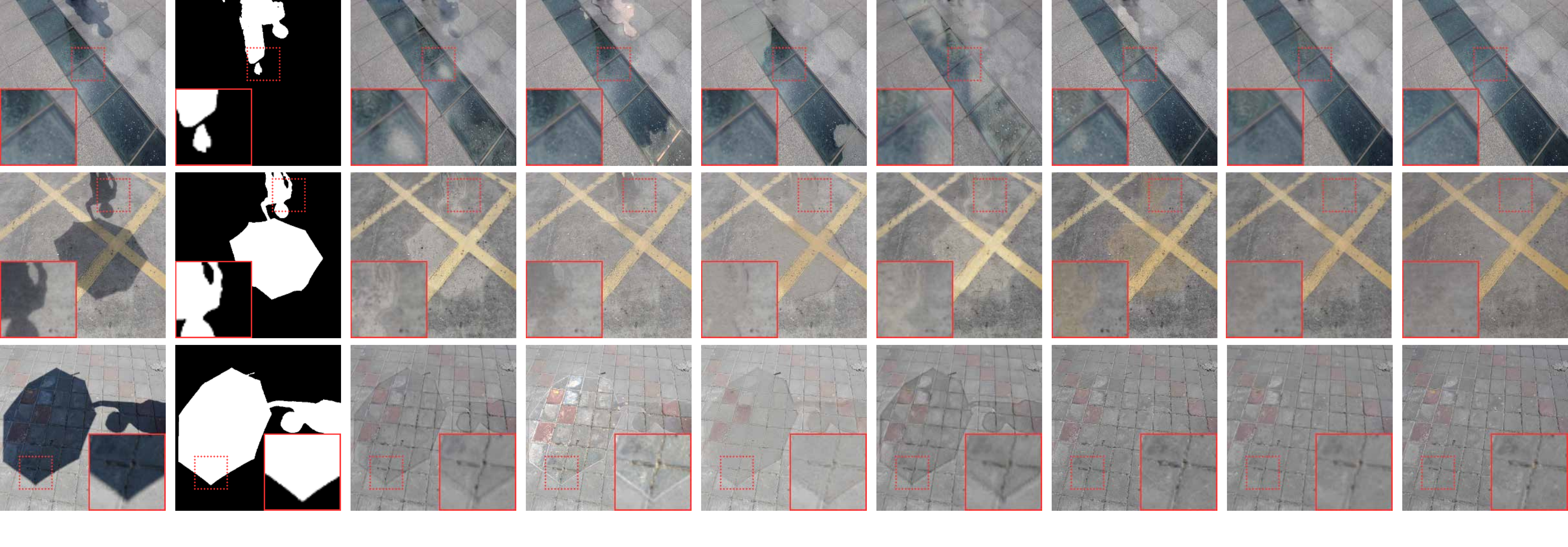}
 \put(0.5,0){\footnotesize Input Shadow}
 \put(11.6,0){\footnotesize Shadow Mask}
 \put(23,0){\footnotesize MS-GAN \cite{hu2019maskshadowgan}}
\put(33.3,0){\footnotesize P+M+D-Net \cite{le2020from}}
\put(47.3,0){\footnotesize G2R \cite{liu2021from}}
\put(57,0){\footnotesize DC-GAN  \cite{jin2021dc}}
\put(69.5,0){\footnotesize AEF \cite{fu2021auto}}
\put(79.5,0){\footnotesize MStructNet}
\put(93.3,0){\footnotesize GT}
 \end{overpic}
  \vspace{-5mm}
 \caption{Qualitative comparison on the ISTD test set.  Please zoom in to see the details. Refer to the Supplemental for more visual comparisons.}
 \label{fig:istd_comp}
 \vspace{-3mm}
 \end{figure*}

We evaluate four StructNet variants (\ie, StructNet-UNet/-STCGAN/-AEF/-SADC) on ISTD+ and ISTD by comparing them with their original versions and show the results in \tableref{tab:structnet_results}. 
We can see that the proposed StructNet improves all four baselines with a significant margin on RSME in the shadow regions over the two datasets. 
In particular, the RMSE of STCGAN decreases from 9.39 to 6.25 (an improvement of 33.4\%) on ISTD+ and from 10.11 to 7.52 (an improvement of 25.6\%) on ISTD. 
\yu{The other three structure-enhanced methods also show clear performance boosts}.  
\ryn{As StructNet-UNet obtains the best results across all counterparts, for convenience, we refer to it as StructNet in all experiments.}

\subsubsection{Comparisons on benchmarks}
\label{subsubsec:comp_on_benchmark}

\revised{We compare StructNet (\ie, StructNet-UNet) and MStructNet with state-of-the-art methods in three benchamrks, \ie, ISTD, ISTD+ and SRD, and the results are shown in \tableref{tab:istd_comp_sotas}, \tableref{tab:istdPlus_comp_sotas}, and \tableref{tab:srd_comp_sotas}. 
Obviously, the results demonstrate that the proposed StructNet and MStructNet outperform all baseline methods in the shadow regions, showing the advantages of our structure-informed approach. 
Notably, StructNet obtains $6.33$ in the shadow regions on the ISTD dataset, with an improvement of 16.7\% and 21.6\% over the  BMNet and EMNet. In terms of the perceptual metric, LPIPS, StructNet also outperforms existing methods by a large margin in both ISTD and ISTD+ datasets.  
Meanwhile, our MStructNet achieves the lowest RMSE  in shadow regions and the lowest LPIPS among competing shadow removal methods in the ISTD, ISTD+, as shown in \tableref{tab:istd_comp_sotas}, \tableref{tab:istdPlus_comp_sotas}.  
In particular, MStructNet has 13.1\% lower RMSE in the shadow region and 31.6\% lower LPIPS on the ISTD+ dataset compared to BMNet.  
In comparison to our StructNet, MStructNet achieves better performance, with a lower RMSE in the non-shadow regions (4.35 vs. 4.71) and similar results in the shadow regions (6.34 vs. 6.33), resulting in a better overall performance (4.68 vs. 4.98) in  ``All'' on the ISTD dataset. 
On the SRD dataset (see \tableref{tab:srd_comp_sotas}), our StructNet and MStructNet obtain the lowest RMSE results in the shadow region, and the best LPIPS perceptual quality assessments.}

\revised{\textbf{Efficiency Comparisons.} In \tableref{tab:time_comp}, we elucidate the efficiency comparisons, encompassing model parameters, floating-point operations (FLOPs), and inference time. StructNet, distinguished by its two-stage shadow removal process, exhibits slightly more parameters relative to other methods, such as DHAN~\cite{cun2020towards} and G2R~\cite{liu2021from}. Although BMNet~\cite{zhu2022bijective} operates with fewer parameters, its high-resolution processing substantially elongates the computational time, rendering it nearly ten times slower than our method. In a striking contrast, StructNet surpasses SP+M+I-Net~\cite{le2021physics}, being 20 times swifter and requiring only one-fourth the FLOPs. Emphasizing the nuanced design of the first-stage network, StructNet intricately embeds MSFE and MFRA modules within each encoder layer.  
This leads to the increasing complexity and results in 6.32G more FLOPs than SGNet~\cite{wan2022style}.  
Our MStructNet improves StructNet by directly processing shadow structure at the feature level, extracting and utilizing shadow-free structure information, thereby reducing the computational overhead.
}

We also display the visual comparison in \figref{fig:istd_comp}. 
The proposed MStructNet can effectively complement low-level cues by integrating multi-level shadow-free structure features, thus facilitating the maximum restoration of the original colors in the umbra and penumbra regions. 
In contrast, other methods either fail to restore the original colors (\eg, MS-GAN and DC-SGAN) or cause obvious artifacts around the penumbra (\eg, G2R and  Param+M+D-Net).

\subsection{\ryn{Evaluation} of StructNet}
\label{subsec:structnet_experiments}

\subsubsection{Effectiveness of \ryn{the MSFE Module}}
\label{subsubsec:effective_sfsc}

\begin{table}[!t]

\caption{Quantitative comparison with the SOTA methods on the ISTD+ dataset. `-' indicates values that are not available.	The best results are highlighted in bold. }
\begin{center}
\footnotesize
		\renewcommand\tabcolsep{8pt}
		\renewcommand\arraystretch{1.}
            \begin{tabular}{l|ccc|c}
            \toprule
         Method $\backslash$ RMSE~$\downarrow$ & S. & N.S. & All& \revised{LPIPS $\downarrow$} \\   
         \midrule

         Guo~\textit{et al.}~\cite{guo2012paired}  &22.0 & 3.1 & 6.1&- \\
         Gong~\textit{et al.}~\cite{gong2016interactive} & 13.3 & - & - &-\\
          SP+M-Net~\cite{le2019shadow} & 7.9 & 3.1 & 3.9 &-\\
          Param+M+D-Net~\cite{le2020from} & 9.7 & 3.0 & 4.0&0.098 \\
          G2R~\cite{liu2021from}  &7.3 & 2.9 & 3.6 &0.092\\
          DC-GAN~\cite{jin2021dc}  &10.3 & 3.5 & 4.6&0.111 \\
          SP+M+I-Net~\cite{le2021physics}  &6.0 & 3.1 & 3.6 &0.080 \\
          BMNet \cite{zhu2022bijective}  & 6.1 & 2.9 & 3.5 &0.079 \\
          SGNet \cite{wan2022style}  & 5.9 & 2.9 & 3.4 &0.091 \\

          \midrule
          StructNet & \textbf{5.3} & \textbf{2.5} & \textbf{3.0} &0.065 \\
          MStructNet &\textbf{5.3} & 2.7 & 3.1 &\textbf{0.054}\\

         \bottomrule
            \end{tabular}
    \end{center}
\label{tab:istdPlus_comp_sotas}
\end{table}

 \begin{table*}[t!]
\caption{\revised{Comparisons of parameters, FLOPS, and network inference time.}}
\begin{center}
    \footnotesize
    \renewcommand\tabcolsep{2.2pt}
    \renewcommand\arraystretch{1.1}
    \begin{tabular}{l|c|c|c|c|c|c|c|c|c}
    \toprule
 Methods   & DHAN~\cite{cun2020towards}&G2R~\cite{liu2021from}  & AEF~\cite{fu2021auto} & DC-GAN~\cite{jin2021dc}& SP+M+I-Net~\cite{le2019shadow} &  BMNet~\cite{zhu2022bijective}   & SGNet~\cite{wan2022style}  & \textbf{StructNet} &\textbf{MStructNet}\\
     \midrule
     Params. (MB) &21.75&22.76 &143.01 &21.16&141.18   &0.37&6.17&67.06&20.62 \\
     FLOPs (G)        &262.87&113.87  &160.32& 105.00& 160.10   &10.99&39.63& 45.95&28.77 \\
     Time (ms)     &41&59  &23   & 6& 60 &33&27 &3.3&2.8 \\
     \bottomrule
\end{tabular}
\label{tab:time_comp}
\end{center}
\end{table*}

\begin{table}[!t]
	\caption{Quantitative comparison with the SOTA methods on the SRD dataset. `-' indicates values that are not available. The best results are highlighted in bold.}
	\begin{center}
        \footnotesize
		\renewcommand\tabcolsep{8pt}
		\renewcommand\arraystretch{1.}
		\begin{tabular}{l|ccc|c}
			\toprule
			Method $\backslash$ RMSE~$\downarrow$ & S. & N.S. & All & \revised{LPIPS $\downarrow$}\\   
			\midrule
                Guo \textit{et al.}~\cite{guo2012paired}  &29.89 & 6.47 & 12.60&- \\
                DeShadowNet~\cite{qu2017deshadownet}  &11.78 & 4.84 & 6.64 &-\\
                DSC~\cite{hu2019direction}  &10.89 & 4.99 & 6.23&0.248 \\
			MS-GAN \cite{hu2019maskshadowgan} & - &-& 7.32&-\\
			AR-GAN \cite{ding2019argan} &7.24&4.71&5.74&-\\
			DHAN \cite{cun2020towards}  &8.39 & 4.67 & 5.46&0.197 \\
			RIS \cite{zhang2020ris}&8.22& 6.05 & 6.78 &-\\
			CLA \cite{zhang2020cla}&8.10 & 6.01 & 6.59 &-\\
			DC-GAN~\cite{jin2021dc} & 7.70 & 3.39 &4.66 &0.109  \\
			CANet \cite{chen2021canet} & 7.82 & 5.88 & 5.98 &-\\
			BMNet \cite{zhu2022bijective}   & 6.96 & \textbf{3.13} &\textbf{4.18}&0.099  \\
			EMNet \cite{zhu2022efficient}& 7.44 & 3.74 & 4.79 & 0.285 \\
			\midrule
			StructNet & 6.93 & 3.94 & 4.81 &0.092\\
			MStructNet  & \textbf{6.69} & 4.28 & 4.97 &\textbf{0.091}  \\

			\bottomrule
		\end{tabular}
	\end{center}
	\label{tab:srd_comp_sotas}
\end{table}

\begin{table}[!ht]
	\caption{Comparison between StructNet variants. The comparisons are conducted  on the ISTD+ dataset from two aspects, \ie, structure-level and image-level shadow removal.
    We denote all variants with $\text{StructNet}(\text{Factor1},\text{Factor2},\text{Factor3})$ where `Factor1' represents the function for the $\text{Bridge}(\cdot)$, `Factor2' means the positions to embed the `Factor1', and `Factor3' is the function for the $\text{Fusion}(\cdot)$ in Eq.\,\ref{eq:structnet_conv}. \revised{``Add'' and ``CONV'' refer to the additive fusion and convolution operations. } }
	\begin{center}
		\renewcommand\tabcolsep{3.pt}
		\renewcommand\arraystretch{1.}
            \begin{tabular}{l|ccc|ccc}
            \toprule
             \multirow{2}{*}{Methods for the first stage $\backslash$ RMSE~$\downarrow$}  & \multicolumn{3}{c|}{Structure-level} & \multicolumn{3}{c}{Image-level } \\
              & S. & N.S. & All & S. & N.S. & All \\
            \midrule
          \makecell[l]{Two-stage shadow removal\\ in \tableref{tab:selection_of_structure_levels} with $l=0.015$} & 5.54 &  1.86 & 2.46 & 5.89 & 2.49 & 3.05 \\
          \midrule
          $\text{StructNet}(\text{MSFE},1,\text{Add})$      & 5.10 & 1.87 & 2.41 & 5.72	& 2.59 & 3.10 \\
          $\text{StructNet}(\text{MSFE},2,\text{Add})$      & 4.80 & 1.88 & 2.36 & 5.55	& 2.59 & 3.07 \\
          $\text{StructNet}(\text{MSFE},3,\text{Add})$     & 4.76 & 2.01 & 2.46 & 5.56	& 2.58 & 3.07 \\
          $\text{StructNet}(\text{MSFE},4,\text{Add})$      & 4.73 & 2.04 & 2.49 & 5.62	& 2.61 & 3.11 \\
          $\text{StructNet}(\text{MSFE},5,\text{Add})$     & 4.74 & 1.97 &	2.43 & 5.55	& 2.57 & 3.06 \\ 
          
          \midrule
          $\text{StructNet}(\text{MSFE},1,\text{MFRA})$       & 4.82 & 1.88 &	2.36 & 5.68	& 2.59 & 3.09 \\
          $\text{StructNet}(\text{MSFE},2,\text{MFRA})$    & 4.32 & 1.92 & 2.31 & 5.39	& 2.59 & 3.05 \\
          $\text{StructNet}(\text{MSFE},3,\text{MFRA})$     & 4.55 & 2.01 & 2.43 & 5.54	& 2.58 & 3.07 \\
          $\text{StructNet}(\text{MSFE},4,\text{MFRA})$ & 4.65 & 1.88 & 2.33 & 5.57 & 2.52 & 3.02 \\
          $\text{StructNet}(\text{MSFE},5,\text{MFRA})$ & 4.58 & 1.92 & 2.35 & 5.43 & 2.52 & 3.00 \\ 
          
          \midrule

          $\text{StructNet}(\text{MSFE},(1\cdots5),\text{Add})$  & 4.82 & 1.87 &	2.35 & 5.50	& 2.60 & 3.07 \\
          $\text{StructNet}(\text{MSFE},(1\cdots5),\text{MFRA})$  & 4.20 & 1.71 &	2.12 & 5.31	& 2.52 & 2.97 \\
          \bottomrule
          \end{tabular}
    \end{center}
	\label{tab:structurenet_design}
\end{table}

We construct different StructNet variants by using different structure-level shadow removal networks and then evaluate the quality of the restored structures (\ie, structure-level RMSEs) from the first stage as well as the quality of the restored images (\ie, image-level RMSEs) from the second stage.

\textbf{Adding MSFE to \ryn{one} single convolution layer}. 
    To avoid the influence of the \ryn{fusion function carried out by the MFRA module, we replace it with a naive element-wise additive operation instead}.
    \ryn{We use $\text{StructNet}(\text{MSFE},j,\text{Add})$ to denote the StructNet whose first-stage network uses the MSFE as Bridge() at the $j$th layer and  the element-wise additive operation as Fusion().} 
    We then obtain five variants, \ie,  $\{\text{StructNet}(\text{MSFE},j,\text{Add})|j\in\{1,2,3,4,5\}\}$.
    \tableref{tab:structurenet_design} shows the results and we  observe: \ding{182} Compared with the naive two-stage shadow removal method  (\ie, vanilla UNet), StructNets with a single MSFE achieves lower structure-level and image-level RMSEs (\ie, $\text{StructNet}(\text{MSFE},1/2/3/4/5,\text{Add})$ in \tableref{tab:structurenet_design}~vs.~two-stage shadow removal in \tableref{tab:selection_of_structure_levels}) in the shadow regions, which demonstrates that the MSFE does benefit the structure-level shadow removal and enhance the image-level shadow removal. 
    \ding{183} In general, if we embed MSFE in a deeper convolution layer, we get lower RMSEs in the shadow regions while slightly higher RMSEs in the non-shadow regions at the structure level. 
    For example, the structure-level RMSE of the shadow region decreases from 5.10 to 4.73 if we add MSFE from the $1$st  to the $5$th layers. 
    We have similar observations on the image-level RMSEs.

\textbf{Adding MSFE to all convolution layers}. 
    We further add MSFE to all layers and denote this variant as $\text{StructNet}(\text{MSFE},(1\cdots 5),\text{Add})$. 
    Compared with $\text{StructNet}(\text{MSFE},5,\text{Add})$, $\text{StructNet}(\text{MSFE},(1\cdots 5),\text{Add})$ has a lower structure-level RMSE (\ie, 1.87) in the non-shadow regions but a slightly higher structure-level RMSE (\ie, 4.82) in the shadow regions. The overall RMSE becomes 2.35, which is smaller than that of $\text{StructNet}(\text{MSFE},5,\text{Add})$.
    In contrast, compared with $\text{StructNet}(\text{MSFE},1,\text{Add})$, $\text{StructNet}(\text{MSFE},(1\cdots 5),\text{Add})$ has a much lower structure-level RMSE in the shadow regions and the same RSME in the non-shadow regions.
    Such observations imply that equipping more convolutions with MSFE can balance the restoration in the shadow and non-shadow regions.
    
\revised{ \textbf{Comparison with other representative convolutions.} We extend our comparison of the MSFE to include three prominent convolutional structures: convolution with a skip function, partial convolution \cite{liu2018image}, and gated convolution \cite{yu2019free}.
    \ding{182} We consider the convolution with a skip function and implement a variant denoted as $\text{StructNet}(\text{CONVSkip})$. In this variant, the fusion function is formulated as an additive operation, and the bridge is designed as a convolutional layer, akin to the convolutional skip connection used in residual networks \cite{he2016deep}.
    \ding{183} In terms of the partial convolution, we follow the way in \cite{liu2018image} and reformulate \reqref{eq:structnet_conv} and \reqref{eq:structnet_conv_res} as $\mathbf{X}_\text{out}^j = \mathbf{B}^j$ and $ \mathbf{B}^j = \text{Bridge}(\mathbf{X}_{\text{in}},\mathbf{M}_\text{in}^j) = \alpha_\mathbf{p}\sum_{\mathbf{q}\in\mathcal{N}_\mathbf{p}}\mathbf{X}_{\text{in}}[\mathbf{q}](1-\mathbf{M}_\text{in}^j[\mathbf{q}])\mathbf{W'}^j_\text{B}[\mathbf{q-p}]$, respectively. We denote the method as $\text{StructNet}(\text{PartialCONV})$. 
    \ding{184} In terms of the gated convolution, we follow \cite{yu2019free} and reformulate \reqref{eq:structnet_conv} and \reqref{eq:structnet_conv_res} as $\mathbf{X}_\text{out}^j = \mathbf{B}^j \odot (\mathbf{X}_{\text{in}}^{j} * \mathbf{W}^{j})$ and $\mathbf{B}^j = \text{Sigmoid}(\mathbf{X}_{\text{in}}^{j} * \mathbf{W}_{\text{f}}^{j})$. We denote the method as $\text{StructNet}(\text{GatedCONV})$.
}

\revised{  We present the comparative results in \tableref{tab:comparison_partial} and identify the following key observations:
    \ding{182} Our method, leveraging MSFE, surpasses all three baseline methods within shadow and non-shadow regions at both the structural and image levels, thereby substantiating the benefits of MSFE.
    \ding{183} The variant $\text{StructNet}(\text{CONVSkip})$ employing a convolutional skip function yields lower RMSEs than the naive two-stage shadow removal method within shadow regions but exhibits higher RMSEs in non-shadow areas. This dichotomy illustrates its potency in shadow removal but underlines an adverse impact on non-shadow regions. This discrepancy occurs primarily because the convolution for the bridge function and the element-wise skip function for fusion inadequately address the shift between shadow and non-shadow regions, failing to overcome the constraints of standard convolution.
    \ding{184} The approach $\text{StructNet}(\text{PartialCONV})$ registers considerably higher RMSEs compared to the naive two-stage method at both structural and image levels. The root cause of this deterioration is the total disregard of the original information within the shadow regions, culminating in marked performance degradation.
    \ding{185} Lastly, $\text{StructNet}(\text{GatedCONV})$ achieves lower RMSEs than the naive two-stage method at the structural level, but incurs higher RMSEs at the image level. This pattern corroborates the capability of gated convolution to restore structural information effectively, while also highlighting its failure to recover finer details.
}

\begin{table}[t]
\caption{\revised{Comparing StructNet variants with MSFE, partial convolution, and gated convolution.}}
	\begin{center}
		\renewcommand\tabcolsep{3.pt}
		\renewcommand\arraystretch{1.}
            \begin{tabular}{l|ccc|ccc}
            \toprule
             \multirow{2}{*}{Methods for the first stage $\backslash$ RMSE~$\downarrow$}  & \multicolumn{3}{c|}{Structure-level} & \multicolumn{3}{c}{Image-level } \\
              & S. & N.S. & All & S. & N.S. & All \\
            \midrule
          \makecell[l]{Two-stage shadow removal\\ in \tableref{tab:selection_of_structure_levels} with $l=0.015$} & 5.54 &  1.86 & 2.46 & 5.89 & 2.49 & 3.05 \\ 
          \midrule
          $\text{StructNet}(\text{CONVSkip})$ & 5.10 & 1.97 & 2.48 & 5.82 & 2.61 & 3.14 \\
          $\text{StructNet}(\text{PartialCONV})$ & 7.99 & 1.99 & 2.89 & 6.50 & 2.57 & 3.22  \\
          $\text{StructNet}(\text{GatedCONV})$ & 5.46 & 1.98 & 2.55 & 6.27 & 2.55 & 3.16 \\
          $\text{StructNet}(\text{MSFE})$  & 4.20 & 1.71 &	2.12 & 5.31	& 2.52 & 2.97 \\

          \bottomrule
          \end{tabular}
    \end{center}
	\label{tab:comparison_partial}
\end{table}

\textbf{Feature comparison.} 
We also compare the proposed MSFE with the standard convolution in \figref{fig:convdiff} by showing their processed features (See \figref{fig:convdiff}(e) vs. (c)). 
Clearly, the visual feature differences between shadow and non-shadow regions of our StructNet are much smaller than those of the naive UNet. 
In addition, as depicted in \figref{fig:convdiff}(f), the proposed StructNet presents much smaller absolute \yu{feature} differences, which also demonstrates its effectiveness.

\subsubsection{Effectiveness of \ryn{the MFRA Module}}
\label{subsubsec:effective_msfra}

\textbf{Adding MFRA to the convolution in the MSFE-based Network.} 
We replace the element-wise additive fusion (\ie, ``Add'') of the variants in \tableref{tab:structurenet_design} (\ie, $\text{StructNet}(\text{MSFE},\star,\text{Add})$) with the proposed MFRA to obtain new variants, $\text{StructNet}(\text{MSFE},\star,\text{MFRA})$, where `$\star$' denotes specific layer indexes used by StructNet. 
We have the following observations: \ding{182} All single-MSFE-based variants with MFRA (\ie, $\text{StructNet}(\text{MSFE},\star,\text{MFRA})$) outperform the variants with the element-wise \ryn{addition} operation, which demonstrates that the proposed aggregation function does enhance shadow removal significantly. For example, $\text{StructNet}(\text{MSFE},2,\text{MFRA})$ achieves 4.32 structure-level RMSE in the shadow regions, outperforming $\text{StructNet}(\text{MSFE},2,\text{Add})$ by 10.0\%.  
\ding{183} When we embed MFRA to all convolutions with MSFE, we find that $\text{StructNet}(\text{MSFE},(1\cdots5),\text{MFRA})$ achieves much better restoration quality in both shadow and non-shadow regions than $\text{StructNet}(\text{MSFE},(1\cdots5),\text{Add})$.

\textbf{Comparison with alternative fusion solutions.} 
We further compare the proposed MFRA with three potential fusion approaches to validate its advantages and effectiveness by comparing the structure restoration quality (\ie, structure-level RMSE). 
\textit{First}, we substitute MFRA with the  ASPP \cite{chen2017rethinking} 
and denote this variant as $\text{StructNet}(\text{MSFE},(1\cdots5),\text{ASPP})$.
\textit{Second}, we degrade MFRA by removing the dynamic weights $\mathbf{B}_s^j$ in Eq.\,\ref{eq:msfra_comb} and adding different scale features directly. We denote this variant as $\text{StructNet}(\text{MSFE},(1\cdots5),\text{MFRA}_\text{v1})$.
\textit{Third}, we construct a degraded variant of MFRA to calculate all four scale features in \secref{subsec:msfra} through Eq.\,\ref{eq:msfra_atrousconv} directly, and we name it as $\text{StructNet}(\text{MSFE},(1\cdots5),\text{MFRA}_\text{v2})$.

We report the comparison results in \tableref{tab:msfra_ablation} and  have the following conclusion: 
\ding{182} Compared with the baseline fusion strategy $\text{StructNet}(\text{MSFE},(1\cdots5),\text{Add})$, $\text{StructNet}(\text{MSFE},(1\cdots5),\text{ASPP})$ obtains a larger structure-level RMSE in the shadow regions, which \ryn{implies} that naively using ASPP is not good enough to fuse multi-scale features for shadow removal. 
\ding{183} Compared with the degraded version $\text{StructNet}(\text{MSFE},(1\cdots5),\text{MFRA}_\text{v1})$, $\text{StructNet}(\text{MSFE},(1\cdots5),\text{MFRA})$ obtains lower RMSEs in both shadow and non-shadow regions, leading to a lower RMSE in ``All'' (\ie, 2.12 vs. 2.26), which demonstrates that combining multi-scale features with dynamically predictive weights via Eq.\,\ref{eq:msfra_comb} indeed helps restore the structure better. 
\ding{184} Using our proposed strategy for extracting the smallest scale features prevents heavy information loss during down-sampling, as shown by the lower RMSEs in shadow and non-shadow regions (4.20 to 4.42 and 1.71 to 1.82, respectively).

\begin{table}[t!]
	\caption{Ablation study on the proposed MFRA module. $\text{StructNet}(\text{MSFE},(1\cdots5),\text{MFRA}_\text{v1})$ is the degraded MFRA by removing the dynamic weights $\mathbf{B}_s^j$ in Eq.\,\ref{eq:msfra_comb} and adding different scale features directly.
    We include another degraded variant of MFRA (\ie, $\text{StructNet}(\text{MSFE},(1\cdots5),\text{MFRA}_\text{v2})$) by computing all four scale features through Eq.\,\ref{eq:msfra_atrousconv} directly.
	}
	\begin{center}
		\renewcommand\tabcolsep{5.0pt}
		\renewcommand\arraystretch{1.}
		\begin{tabular}{l|ccc}
			\toprule
			Variants $\backslash$ Structure-level RMSE~$\downarrow$ & S. & N.S. & All \\   
			\midrule
			$\text{StructNet}(\text{MSFE},(1\cdots5),\text{Add})$  & 4.82 & 1.87 & 2.35 \\
			$\text{StructNet}(\text{MSFE},(1\cdots5),\text{ASPP})$  &5.15 & 1.81 & 2.37 \\
			$\text{StructNet}(\text{MSFE},(1\cdots5),\text{MFRA}_\text{v1})$    &4.65 & 1.79 & 2.26 \\
			$\text{StructNet}(\text{MSFE},(1\cdots5),\text{MFRA}_\text{v2})$ &4.42 & 1.82 & 2.25 \\
			\midrule
			$\text{StructNet}(\text{MSFE},(1\cdots5),\text{MFRA})$    &4.20 & 1.71 & 2.12 \\
			\bottomrule
		\end{tabular}
	\end{center}
	\label{tab:msfra_ablation}
\end{table}

\subsection{Effectiveness of MStructNet}
\label{subsec:mstructnet_experiments}

\begin{table}[t]
	\caption{Ablation experiment of MStructNet on the ISTD+ dataset, with respect to different structure level utilization.}
	\begin{center}
	\renewcommand\tabcolsep{5pt}
	\renewcommand\arraystretch{1.}
	\begin{tabular}{cccc|ccc}
		\toprule

       \multicolumn{4}{c|}{Structure levels} & \multicolumn{3}{c}{Image-level RMSE~$\downarrow$} \\
		0.005&0.015 & 0.045 & 0.1& S. & N.S. & All   \\
		\midrule

		 \checkmark & & & &5.46&2.90 & 3.32 \\
		  &\checkmark & & &5.46&2.78 & 3.22 \\
		  & &\checkmark & &5.61&2.73 & 3.20 \\ 
		  & & &\checkmark &5.66&2.74 & 3.21 \\ 
		 \midrule
		  \checkmark&\checkmark & & &5.46&2.81 & 3.24 \\ 
		  \checkmark&\checkmark &\checkmark & &5.42&2.79  & 3.22 \\ 
		  \checkmark&\checkmark &\checkmark &\checkmark &5.29&2.73 & 3.15  \\

		\bottomrule
	\end{tabular}
\end{center}
\label{tab:mstructurenet_design}
\end{table}

\textbf{Numbers of Structure Levels.} 
In \secref{subsec:mstructnet_pipeline}, the structure-aware encoder \ryn{ is made up of several blocks, with each block representing one structure level and containing two convolution layers equipped with the MSFE and MFRA modules. Note that we set two convolution layers for each block due to the empirical results in \tableref{tab:structurenet_design} (as  $\text{StructNet}(\text{MSFE},2,\text{Add/MFRA})$ achieves the lowest RMSE in ``All'', among  single-convolution based variants)}.

\textbf{Number of blocks (or structure levels) in MStructNet.}
As \ryn{discussed} in \secref{subsec:mstructnet_pipeline}, \ryn{each block contains MSFE and MFRA modules to form a} structure level, and \yu{\ryn{the final MStructNet fuses structures of all different levels}.}  
Here, we study the effects of using different \ryn{numbers of blocks in the structure-aware encoder} to validate the advantages of \yu{exploiting} \ryn{the multi-level structures}. 
Specifically, we may obtain four variants of MStructNet by using a single structure selected from $\{0.005, 0.015, 0.045, 0.1\}$, \ryn{and are} denoted as: $\{\text{MStructNet}(l)|l\in\{0.005, 0.015, 0.045, 0.1\}\}$. 
\ryn{We then} gradually add more structures to MStructNet$(0.005)$ to obtain three more variants, denoted as: \yu{MStructNet$(\{0.005,0.015\})$, MStructNet$(\{0.005,0.015,0.045\})$, and MStructNet$(\{0.005,$ $0.015,0.045,0.1\})$, respectively.} The last version denotes the final version of $\text{MStructNet}$.
As reported in \tableref{tab:mstructurenet_design}, we can see that: \ding{182} MStructNet with the structure level 0.015 shows the best results among all single structure level \ryn{variants}. \ding{183} when we add more structure levels, the restoration quality gradually improves and MStructNet with all four structure levels achieves the lowest RMSE in the shadow and non-shadow regions, which  confirms that the utilization of  multi-level \yu{non-shadow} structures at the feature level can indeed benefit the image-level shadow removal.

\begin{table}[t!]
\caption{\revised{Comparisons of different loss ratios for perceptual loss. The $\lambda_{1}= 1$ by default. The MStructNet model and the ISTD+ dataset were utilized for these experiments.}}
\begin{center}
    \footnotesize
    \renewcommand\tabcolsep{1.5pt}
    \renewcommand\arraystretch{1.1}
    \begin{tabular}{l|c|c|c|c|c|c|c|c|c|c}
    \toprule
         \multirow{2}{*}{$\lambda_{2}$ ratio}  &
         \multicolumn{3}{c|}{RMSE~$\downarrow$} & \multicolumn{3}{c|}{PSNR~$\uparrow$} & \multicolumn{3}{c|}{SSIM~$\uparrow$} & \multirow{2}{*}{\revised{LPIPS $\downarrow$}} \\
         & S. & N.S. & All & S. & N.S. & All & S. & N.S. & All \\
     \midrule 
      0.01 & 5.45 & 2.66 & 3.12 & 38.05 & 36.00 & 33.39 & 0.990 & 0.976 & 0.960 & 0.062\\
      \midrule
      0.1 (Ours) & \textbf{5.34} & 2.73 & \textbf{3.10} & \textbf{38.27} & 35.83 & 33.34 & \textbf{0.990} & 0.976 & 0.962 & \textbf{0.054}\\
      \midrule 
      1 & 5.55 & 2.73 & 3.19 & 38.00 & 36.46 & 33.57 & 0.990 & 0.977 & 0.963 & 0.059\\
      \midrule
      10 & 5.64 & 2.79 & 3.26 & 38.03 & 36.61 & 33.68 & 0.990 & 0.977 & 0.962 & 0.059\\
      \bottomrule
\end{tabular}
\label{tab:loss_comp}
\end{center}
\end{table}

\revised{\textbf{Different loss ratio $\mathit{L}_{perc}$ in Eq. \eqref{eq:loss_overall}}. 
We have also conducted experiments using different combinations of loss ratios, \ie,  $\lambda_{2}\in \{0.01, 0.1, 1, 10\}$, and the results are presented in \tableref{tab:loss_comp}.
It is worth noting that the ratio combination of $\lambda_{1}=1$ and $\lambda_{2}=0.1$ achieves the best results in all metrics in the shadow regions. 
Note that we have not conducted meticulous adjustments to the loss ratios.
}

\section{Conclusion}
In this paper, we have systematically investigated the utilization and efficacy of image structure for single-image shadow removal.
\textit{First}, we have built vanilla UNet-based networks to restore the shadow-free structure of the input shadow image, and revealed that image structure can help enhance the quality of shadow-removed images significantly. 
\textit{Second}, we have proposed a novel two-stage removal network named structure-informed shadow removal network (StructNet).
It includes two new modules for the utilization of structure information, \ie, \textit{mask-guided shadow-free extraction (MSFE) module} and \textit{multi-scale feature \& residual aggregation (MFRA) module}, to extract the image structural features and regularize the feature consistency, respectively. 
We have shown that StructNet can help improve the performances of three state-of-the-art methods.
\textit{Third}, based on StructNet, we have further proposed a self-contained shadow removal method to fully excavate the potential of multi-level structures at the feature level, named \textit{multi-level StructNets (MStructNet)}, which has fewer parameters and low computational costs.
The extensive results on three public datasets have also demonstrated the advantages and effectiveness of the proposed  StructNet and MStructNet.

\bibliographystyle{IEEEtran}
\bibliography{main}

\vspace{-15mm}
\begin{IEEEbiography}[{\includegraphics[width=0.9in,height=1.2in,clip,keepaspectratio]{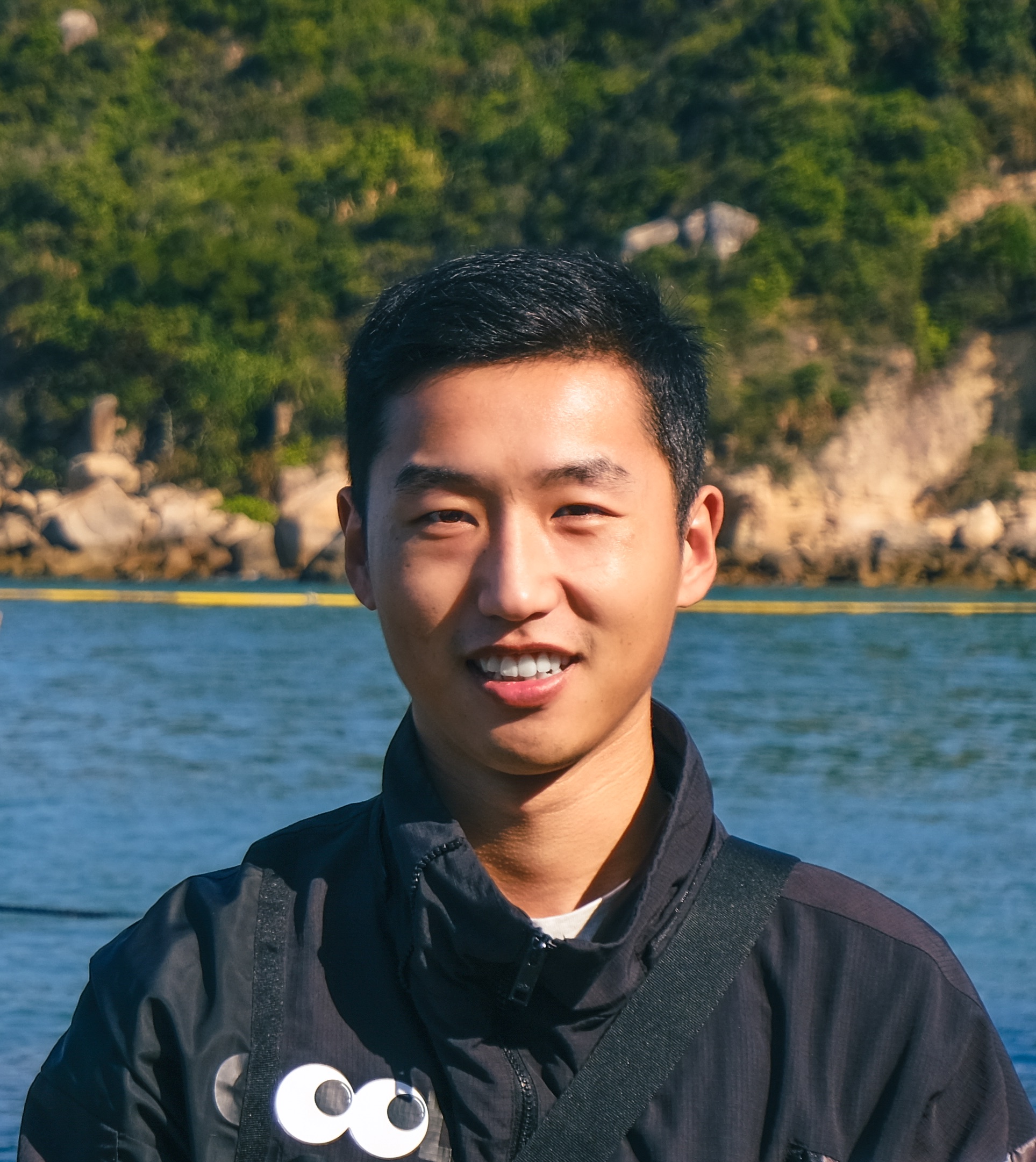}}]{Yuhao Liu} received his B.Eng. degree from Zhengzhou University in 2019 and his M.Sc. degree from Dalian University of Technology in 2022. He is a Ph.D. candidate in Computer Science at the City University of Hong Kong, where his research is focused on solving computer vision and image processing problems.
\end{IEEEbiography}
\vspace{-15mm}
\begin{IEEEbiography}[{\includegraphics[width=0.9in,height=1.2in,clip,keepaspectratio]{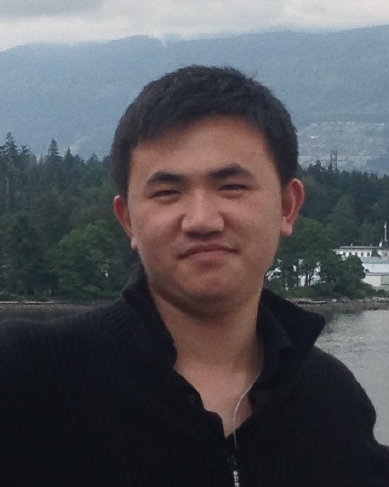}}]{Qing Guo} received the Ph.D. degree in computer application technology from Tianjin University, China. He was a research fellow at the Nanyang Technology University, Singapore, from Dec. 2019 to Aug. 2020 and the Wallenberg-NTU Presidential Postdoctoral Fellow from Sep. 2020 to Sep. 2022. He is currently a senior scientist and PI at the Institute of High Performance
Computing (IHPC) and Center for Frontier AI Research at A*STAR, Singapore. His research interests include computer vision, AI security, and image processing. He is a member of IEEE.
\end{IEEEbiography}
\vspace{-15mm}
\begin{IEEEbiography}[{\includegraphics[width=0.9in,height=1.2in,clip,keepaspectratio]{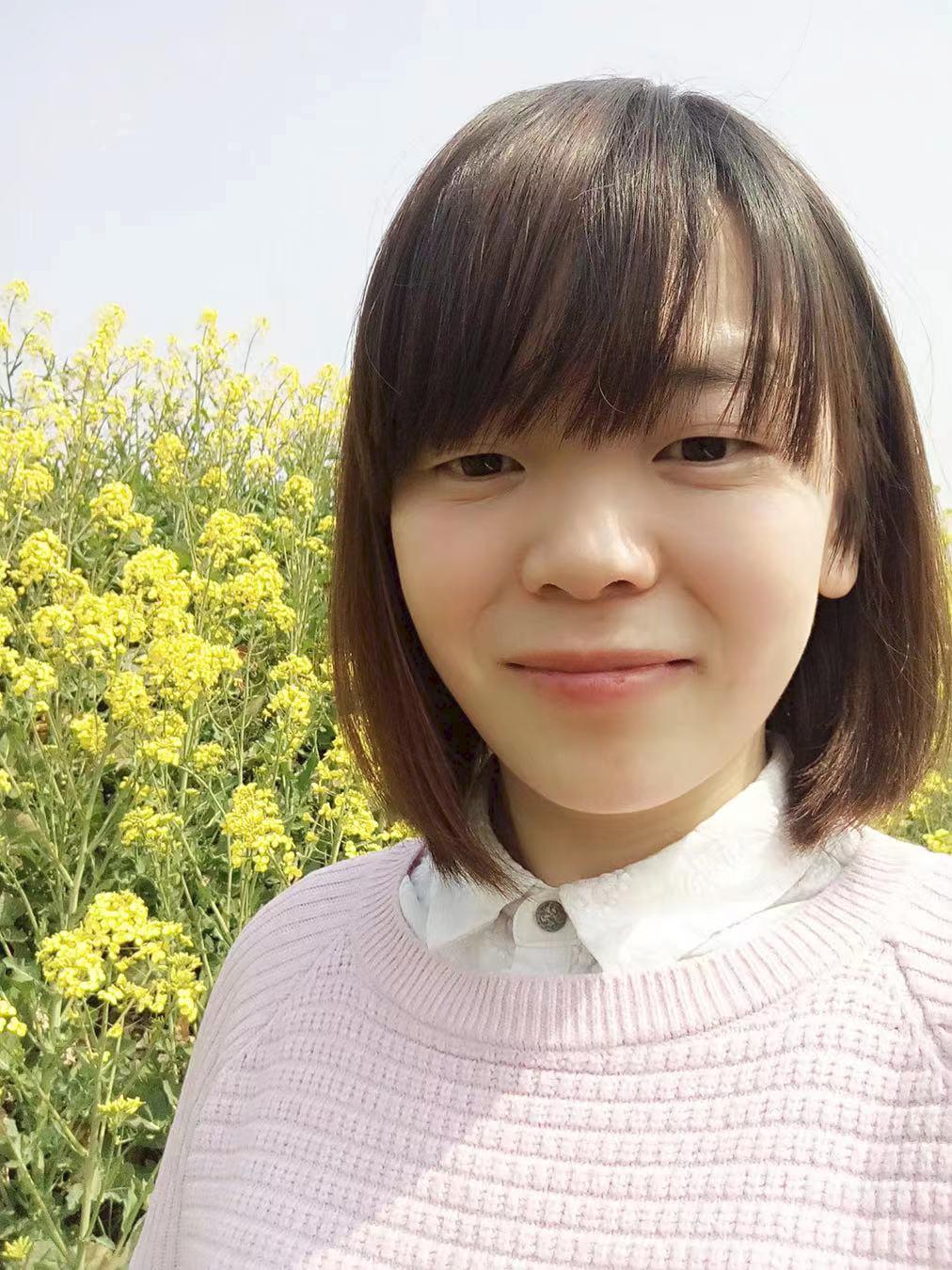}}]{Lan Fu} received a Ph.D. degree in Computer Science and Engineering from the University of South Carolina, Columbia, SC, USA. Prior to that, she received an M.S. degree in Biomedical Engineering from Tianjin University, Tianjin, China. Currently, she is a Senior Research Engineer with InnoPeak Technology Inc., Palo Alto, CA, USA. Her research interests include computer vision, deep learning, and image processing.
\end{IEEEbiography}
\vspace{-15mm}
\begin{IEEEbiography}[{\includegraphics[width=0.9in,height=1.2in,clip,keepaspectratio]{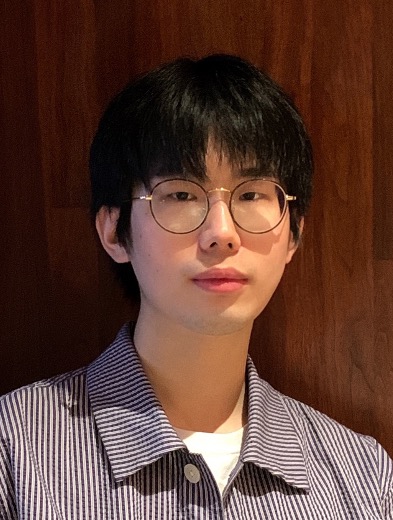}}]{Zhanghan Ke} is a Ph.D. candidate at City University of Hong Kong with a B.Eng. from Northeastern University (China). He serves as a reviewer for several computer vision conferences (\eg, CVPR, ICCV, and ECCV), and journals (\eg, TPAMI, IJCV, and TCSVT). His research interests currently include  semi and self-supervised learning and its applications in computer vision.
\end{IEEEbiography}
\vspace{-10mm}
\begin{IEEEbiography}[{\includegraphics[width=0.9in,height=1.2in,clip,keepaspectratio]{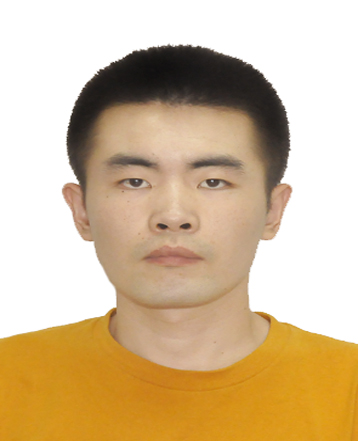}}]{Ke Xu} is currently with the Department of Computer Science at the City University of Hong Kong. He obtains dual Ph.D. degrees from the Dalian University of Technology and the City University of Hong Kong. His research interests include deep learning, object detection, and image enhancement and editing.
\end{IEEEbiography}
\vspace{-5mm}
\begin{IEEEbiography}[{\includegraphics[width=0.9in,height=1.2in,clip,keepaspectratio]{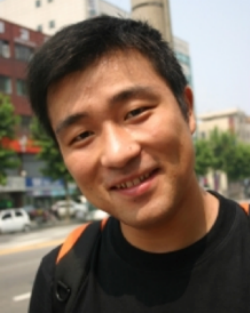}}]{Wei Feng} received a Ph.D. degree in computer science from the City University of Hong Kong in 2008. From 2008 to 2010, he was a research fellow at the Chinese University of Hong Kong and the City University of Hong Kong. He is now a full Professor at the School of Computer Science and Technology, College of Computing and Intelligence, Tianjin University, China. His major research interests are active robotic vision and visual intelligence, specifically including active camera relocalization and lighting recurrence, general Markov Random Fields modeling, energy minimization, active 3D scene perception, SLAM, video analysis, and generic pattern recognition. Recently, he focuses on solving preventive conservation problems of cultural heritages via computer vision and machine learning. He is the Associate Editor of Neurocomputing and the Journal of Ambient Intelligence and Humanized Computing.
\end{IEEEbiography}
\vspace{-5mm}
\begin{IEEEbiography}[{\includegraphics[width=0.9in,height=1.2in,clip,keepaspectratio]{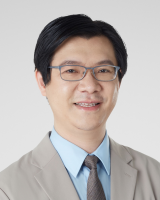}}]{Ivor W. Tsang} is director of A*STAR Centre for Frontier AI Research (CFAR) since Jan 2022. Previously, he was a Professor of Artificial Intelligence at the University of Technology Sydney (UTS), and Research Director of the Australian Artificial Intelligence Institute (AAII). Prof Tsang also serves on the Editorial Board for the Journal of Machine Learning Research, Machine Learning, Journal of Artificial Intelligence Research, IEEE Transactions on Pattern Analysis and Machine Intelligence, IEEE Transactions on Artificial Intelligence, IEEE Transactions on Big Data, and IEEE Transactions on Emerging Topics in Computational Intelligence. He serves as a Senior Area Chair/Area Chair for NeurIPS, ICML, AAAI, and IJCAI, and the steering committee of ACML.
\end{IEEEbiography}
\vspace{-5mm}
\begin{IEEEbiography}[{\includegraphics[width=0.9in,height=1.2in,clip,keepaspectratio]{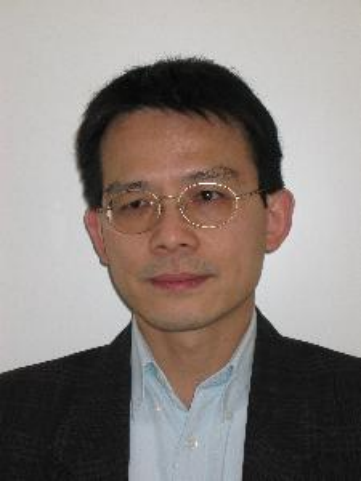}}]{Rynson W.H. Lau} received his Ph.D. degree from the University of Cambridge. He was on the faculty of Durham University and is now with the City University of Hong Kong.
Rynson serves on the Editorial Board of the International Journal of Computer Vision (IJCV) and Computer Graphics Forum. He has served as the Guest Editor of a number of journal special issues, including ACM Trans. on Internet Technology, IEEE Trans. on Multimedia, IEEE Trans. on Visualization and Computer Graphics, and IEEE Computer Graphics \& Applications. He has also served on the committee of a number of conferences, including Program Co-chair of ACM VRST 2004, ACM MTDL 2009, IEEE U-Media 2010, and Conference Co-chair of CASA 2005, ACM VRST 2005, ACM MDI 2009, ACM VRST 2014. Rynson's research interests include computer graphics and computer vision.
\end{IEEEbiography}

\vfill

\end{document}